\documentclass[pdflatex,sn-mathphys-num]{sn-jnl}

\usepackage{graphicx}%
\usepackage{multirow}%
\usepackage{amsmath,amssymb,amsfonts}%
\usepackage{amsthm}%
\usepackage{mathrsfs}%
\usepackage[title]{appendix}%
\usepackage{xcolor}%
\usepackage{textcomp}%
\usepackage{manyfoot}%
\usepackage{booktabs}%
\usepackage{algorithm}%
\usepackage{algorithmicx}%
\usepackage{algpseudocode}%
\usepackage{listings}%

\theoremstyle{thmstyleone}%
\theoremstyle{thmstyletwo}%

\theoremstyle{thmstylethree}%

\begin{document}

\title{Socratic agents for autonomous scientific discovery in high-dimensional physical systems}


\author[1,2]{\fnm{Xianrui} \sur{Zeng}}
\author[1]{\fnm{Pengfei} \sur{Liu}}
\author[3]{\fnm{Yirui} \sur{Zang}}
\author[1]{\fnm{Yang} \sur{Shen}}
\author[1,2]{\fnm{Fei} \sur{Yu}}
\author[1,2]{\fnm{Chunlei} \sur{Yu}}
\author[4]{\fnm{Minghao} \sur{Liu}}
\author*[1]{\fnm{Yang} \sur{Du}}\email{yangdu@ucas.ac.cn}

\affil[1]{\orgdiv{Hangzhou Institute for Advanced Study}, \orgname{University of Chinese Academy of Sciences}, \orgaddress{\city{Hangzhou}, \postcode{310024}, \country{China}}}

\affil[2]{\orgdiv{Shanghai Institute of Optics and Fine Mechanics}, \orgname{Chinese Academy of Sciences}, \orgaddress{\city{Shanghai}, \postcode{201800}, \country{China}}}

\affil[3]{\orgdiv{Department of Thoracic Surgery}, \orgdiv{Shanghai Pulmonary Hospital}, \orgdiv{School of Medicine}, \orgname{Tongji University}, \orgaddress{\city{Shanghai}, \postcode{200433}, \country{China}}}

\affil[4]{\orgdiv{ZODA}}

\abstract{The automation of scientific discovery has reached an inflection point. While artificial-intelligence systems now operate instruments, optimize parameters and generate hypotheses with increasing proficiency, most remain procedural: they execute workflows whose scientific objectives, action spaces and success criteria are fixed by human designers. True autonomous science, by contrast, demands epistemic autonomy---the capacity to construct, challenge and revise physical explanations in response to experimental evidence. Here we introduce AHOIS, a multi-agent AI scientist that embeds Socratic midwifery into closed-loop experimentation as a computational principle for epistemic autonomy. A dedicated physics-critic agent interrogates candidate hypotheses through structured causal questioning, physical-constraint checking, counterexample generation and falsification-criteria formulation, transforming implicit and potentially brittle reasoning into explicit, testable and revisable mechanistic models. We evaluate AHOIS on a real multimode-fibre optical platform, a representative high-dimensional physical system that combines complex wave transformations, indirect detection, environmental drift and multi-modal acquisition. Operating without prior encoding schemes, classifiers or speckle models, the system autonomously proposed and experimentally validated a random-interference encoding hypothesis, discovered task-adaptive sparse-measurement strategies, diagnosed distinct failure modes (encoding instability, fluorescence contamination and detector noise) and translated a published imaging protocol into an executable experimental workflow on a non-original configuration. The discovered encoding yielded 16 $\times$ 16 measurements with an effective rank of 56.9 and classification accuracies of 76.97\% on MNIST and 83.17\% on Fashion-MNIST. Controlled ablations demonstrate that Socratic interrogation substantially improves physical consistency, hypothesis completeness, uncertainty calibration and downstream experimental-plan validity. These results establish a principled route from workflow automation towards evidence-grounded, self-correcting autonomous scientific discovery in complex physical environments.
}

\keywords{autonomous scientific discovery, multi-agent systems, hypothesis-driven experimentation, multimode fibre imaging, Socratic reasoning, trustworthy measurement, AI for science}

\maketitle

\section{Introduction}\label{sec1}

Scientific discovery is not the execution of a fixed sequence of operations, but a continual reorganization of questions, explanations and experiments as evidence accumulates~\cite{king2004functional,king2009automation}. Researchers move repeatedly between observation, abstraction, hypothesis formation, experimental intervention and critical revision. A measurement acquires scientific meaning only when it can support or reject an explanation of the underlying system~\cite{schmidt2009distilling}; similarly, a hypothesis becomes scientifically useful only when its assumptions are explicit, its predictions are testable and its failure conditions are identifiable. Automating scientific discovery therefore demands more than faster data acquisition or parameter optimization~\cite{burger2020mobile,szymanski2023autonomous}. It requires an artificial system capable of maintaining an evolving model of physical reality, identifying what it does not yet understand, and selecting experiments that reduce uncertainty between competing explanations~\cite{gottweis2026accelerating,ghareeb2026multi}.

Recent advances in large language models and agentic systems have begun to automate substantial components of this process. Language models have been used for scientific literature synthesis, hypothesis generation, protocol construction, code development, data analysis and manuscript preparation~\cite{gottweis2026accelerating,ghareeb2026multi,endtoend2024,multiagent2024,wang2024agent,li2024,white2024}. When connected to simulators, robotic laboratories and scientific instruments, agents can additionally execute experimental procedures, optimize control parameters and update decisions from measured outcomes~\cite{endtoend2024,multiagent2024,boiko2023,szymanski2023}. Multi-agent architectures further distribute these functions across specialized roles, allowing literature retrieval, planning, experimentation and analysis to operate within a closed loop~\cite{lyu2026evoscientist,zhang2026agentic,zhu2026evomaster,zhao2026scienceearth,bianchi2026collective}. Emerging frameworks such as EvoMaster~\cite{zhu2026evomaster}, Science Earth~\cite{zhao2026scienceearth} and EvoSci~\cite{xiong2026evosci} envision planet-scale, self-organizing agent ecosystems for open-ended discovery. Aster has demonstrated that iterative program improvement can accelerate discovery by more than an order of magnitude~\cite{bicker2026aster}, while OR-Agent introduces structured tree-based research workflows that explicitly manage branching hypothesis exploration and systematic backtracking~\cite{liu2026oragent}. MatClaw and AutoMOOSE have extended agentic automation to materials exploration and multiphysics simulation, respectively~\cite{zhang2026matclaw,manna2026automoose}. As these systems move from text-based assistance to instrument-connected experimentation, the meaning of autonomy becomes more demanding. In many current demonstrations, the agent operates within a scientific frame already defined by human researchers: the objective, intervention space, evaluation metric and interpretation route are specified in advance. What remains less established is whether an artificial system can infer, from ambiguous experimental evidence, which physical explanation is credible, which assumptions remain uncertain and which intervention would discriminate between competing mechanisms~\cite{bisht2026notbuilt,ho2026soundness,autoresearch2026bench}.

This distinction exposes a deeper challenge for autonomous science. A system may competently operate instruments and still reason from an incorrect physical model; it may produce a fluent hypothesis without exposing its hidden assumptions; and it may interpret an experimental failure as evidence against a theory when the actual cause is calibration drift, detector noise or an execution lapse~\cite{wu2026epistemic,chen2026stateful,swaminathan2026scitrace}. Generic self-reflection or post hoc review is insufficient because it often evaluates linguistic plausibility rather than the causal and evidential structure of a scientific claim~\cite{shinn2023,yao2023}. We argue that progress from procedural autonomy to epistemic autonomy requires an explicit mechanism for interrogating how an agent knows what it claims to know. We therefore introduce Socratic midwifery as a computational principle for autonomous scientific reasoning~\cite{jia2026scimind,oh2026multipersona,li2026causalagent}. Recent multi-agent debate systems have shown that adversarial interrogation can surface hidden assumptions and strengthen hypothesis generation~\cite{jia2026scimind,oh2026multipersona,yang2026causalab}. In the classical sense, the role of Socratic inquiry is not to supply an answer, but to elicit and refine knowledge through disciplined questioning. In an artificial scientist, this principle can be operationalized by requiring proposed explanations to survive clarification of concepts, exposure of assumptions, consistency with physical constraints, causal probing, counterexamples, alternative hypotheses and falsification criteria. The objective is not to imitate philosophical dialogue, but to transform implicit and potentially brittle reasoning into explicit, testable and revisable mechanistic models.

High-dimensional optical systems based on complex media provide a stringent environment in which to test this capability. In scattering media and multimode waveguides, a controllable input field is transformed by multiple scattering, mode mixing and interference into a high-dimensional optical response that can often be described, at least locally, by a transmission matrix~\cite{vellekoop2007,popoff2010measuring,mosk2012,rotter2017light}. This transformation makes complex media scientifically distinctive: the same apparent disorder that degrades conventional imaging can also encode information, enable wavefront control and support computational reconstruction~\cite{bertolotti2012,katz2014,mosk2012}. Multimode fibres are a compact and experimentally accessible representative of this class. Thousands of guided modes propagate with calibration-dependent phases and amplitudes, converting input wavefronts, object interactions and environmental perturbations into speckle-like intensity measurements~\cite{cizmar2012,cao2023controlling,ploschner2015seeing}. Under these conditions, successful operation cannot be reduced to retrieving a known protocol or optimizing a single scalar objective. An autonomous scientific agent must determine whether a measured fluctuation reflects useful physical encoding, uncontrolled decorrelation, detector noise, modality-specific artefacts or execution failure, and must select interventions that separate these competing explanations. Such ambiguity creates precisely the type of epistemic problem that a scientific agent must resolve.

Here we introduce AHOIS, a multi-agent framework for Socratic, physics-grounded autonomous scientific discovery in high-dimensional experimental systems. AHOIS separates scientific hypothesis formation, Socratic physical interrogation, hardware abstraction, system-integrity monitoring and quantitative inference across five interacting agents. Its defining feature is not the division of labour itself, but the epistemic loop between the hypothesis-generating agent and the Socratic physics agent. Candidate explanations are iteratively transformed into structured scientific states comprising the proposed mechanism, physical assumptions, competing interpretations, predicted outcomes, discriminating experiments, uncertainty estimates and rejection conditions. These states are then translated into executable operations on a real multimode-fibre platform, and the resulting measurements are returned to the reasoning loop to revise the system’s model and determine the next experiment.

We assess AHOIS through a sequence of increasingly open-ended experimental problems. The system first constructs a working physical model of a dual-modality multimode-fibre instrument and establishes the constraints required for reliable operation. It then proposes and experimentally evaluates the hypothesis that object-dependent coherent backscattering can function as a high-dimensional random-interference encoder, rather than merely as an imaging nuisance. From subsequent evidence, it identifies task-adaptive sparse-measurement strategies, distinguishes encoding instability from fluorescence and detector-noise failure modes, selects corresponding computational interventions, and transfers a published reconstruction principle to a non-original experimental configuration. Finally, controlled ablations isolate the contribution of Socratic questioning and show that clarification, causal probing and counterexample-based interrogation improve physical consistency, hypothesis completeness, uncertainty calibration and downstream plan validity. Together, these experiments position the multimode-fibre system not as the boundary of the framework, but as a demanding instance of a broader class of high-dimensional physical environments. They demonstrate how structured inquiry can convert an agent from an operator of scientific tools into a system that constructs, challenges and revises experimentally grounded physical explanations.

\section{AHOIS: a Socratic multi-agent architecture for physical experimentation}\label{sec2}

\begin{figure*}[h]
\centering
\includegraphics[width=0.95\textwidth]{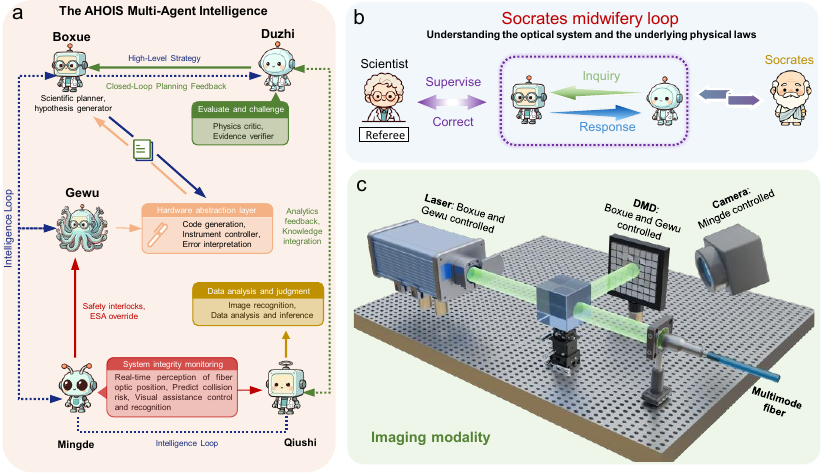}
\caption{The AHOIS multi-agent architecture and the physical testbed. \textbf{a}, Five functionally separated agents cooperate through a shared scientific state. Boxue formulates hypotheses and experimental plans; Gewu translates plans into hardware-executable operations; Mingde monitors system integrity and data validity; Qiushi performs modelling, reconstruction and quantitative analysis; and Duzhi interrogates Boxue's reasoning as a Socratic physics critic. \textbf{b}, Socratic midwifery loop. Duzhi challenges candidate explanations through clarification, physical-constraint checking, causal probing and counterexample generation under human supervision, exposing hidden assumptions and missing causal links without directly prescribing the scientific conclusion. \textbf{c}, High-dimensional multimode-fibre optical platform used as the experimental testbed for agent-controlled wavefront shaping, multimode propagation and image acquisition.}
\label{fig1}
\end{figure*}

AHOIS was designed to separate the intellectual, physical and evaluative components of autonomous experimentation while keeping them coupled through a shared scientific state (Fig.~\ref{fig1}a). At each step, this state records the current hypothesis, its physical assumptions, the proposed experimental action, the expected observation, the relevant uncertainty and the evidence returned by the instrument. The agents therefore do not operate as a fixed sequential pipeline. Instead, they exchange hypotheses, executable plans, integrity checks and analytical results, allowing the research trajectory to be revised as experimental evidence accumulates.

Boxue is the strategy agent. It interprets a high-level research goal, decomposes it into candidate hypotheses and produces machine-readable experimental plans. Gewu is the hardware-abstraction agent. It converts these plans into executable routines for laser operation, digital micromirror device (DMD) pattern generation, camera triggering and metadata recording. Mingde is the system-integrity agent. It audits the live data stream by checking exposure, rejecting saturated frames, validating dark frames, assessing frame completeness and quantifying speckle repeatability. Qiushi is the modelling agent. It implements data loaders, reconstruction models, training procedures, hyperparameter searches and quantitative analyses. Duzhi is the Socratic physics critic. Rather than acting as a second planner, Duzhi is constrained to interrogate Boxue's reasoning by exposing hidden assumptions, missing constraints, unsupported causal links and plausible but incorrect interpretations (Fig.~\ref{fig1}b).

This role separation is central to the design. AHOIS is not implemented as a single monolithic prompt that directly maps a user request to an experimental action~\cite{ouyang2022,wei2022,schick2023,yao2023,shinn2023}. The modular structure makes it possible to inspect each decision, replay the reasoning trajectory and isolate the contribution of Socratic criticism from hardware execution or downstream analysis. In this architecture, questioning is treated as part of the scientific process itself: a candidate explanation must be clarified, physically constrained and made falsifiable before it is converted into an experimental intervention.

The physical testbed is a multimode-fibre optical platform (Fig.~\ref{fig1}c). It consists of a laser source, a DMD for wavefront shaping, coupling optics, a multimode fibre, a sample plane, collection optics and imaging detectors. The platform is high-dimensional in a physical rather than merely digital sense: controllable input fields are mapped through many coupled fibre modes into speckle-like intensity responses, and the measured signal depends jointly on wavefront modulation, multimode propagation, object interaction, collection geometry and detection modality. Far-field holographic imaging and fluorescence imaging therefore share parts of the optical hardware but obey different physical models. This combination of shared hardware, modality-dependent observables and calibration-sensitive transformations provides the experimental setting in which AHOIS must distinguish useful physical responses from artefacts, drift and execution failures.

A human scientist supervises AHOIS at predefined decision points. The supervisor can veto unsafe actions, correct hardware-incompatible plans or annotate uncertain interpretations, while all agent states and experimental outcomes are logged. This supervision is used as a safety and quality gate rather than as a replacement for the agent's planning, execution and analysis loop~\cite{sumers2024,boiko2023,zhu2026lap,swaminathan2026scitrace}.

\section{Autonomous discovery of backscattered speckle encoding}\label{sec3}

We next asked whether AHOIS could use its physical understanding of the multimode-fibre platform to identify a function that had not been specified in advance. The system was given the open-ended prompt: ``Based on your own understanding of the underlying physical principles, what else could this system do?'' No encoding scheme, recognition task, validation metric or target application was provided. The purpose of this experiment was therefore not to execute a predefined imaging protocol, but to test whether the agentic loop could generate a falsifiable physical hypothesis and convert it into an experimental measurement.

Boxue examined the illumination, object-scattering and collection pathways of the platform and noted that the collected backscattered light contained more information than the local reflectance used in conventional raster-scanning imaging. Under coherent illumination, fields scattered from different illuminated regions of a rough object acquire object-dependent amplitudes and phases. After coupling into the collection multimode fibre, these fields undergo mode mixing and coherent superposition before square-law detection. Boxue therefore proposed that the apparently random backscattered speckle should not necessarily be treated as an imaging artefact. Instead, the optical system might act as a physical random-interference encoder, transforming object-dependent scattering into a compact measurement from which object identity could be inferred without reconstructing a visually recognizable image.

Duzhi then challenged whether the proposed speckle response was genuinely information-bearing or merely reflected low-dimensional variations in total collected intensity. This interrogation led Boxue to define three criteria for testing the hypothesis before attempting neural decoding. First, the encoded measurements should not collapse into highly redundant responses across different inputs, which was assessed using a correlation matrix. Second, the measurement ensemble should occupy a multidimensional feature space rather than a few dominant intensity components, which was assessed by singular-value decomposition and effective-rank analysis. Third, the encoded intensities should use a measurable dynamic range without being dominated by saturation, near-zero values or uniform background. These criteria converted the initial physical intuition into experimentally testable conditions for random-interference encoding.

\begin{figure*}[h]
\centering
\includegraphics[width=0.95\textwidth]{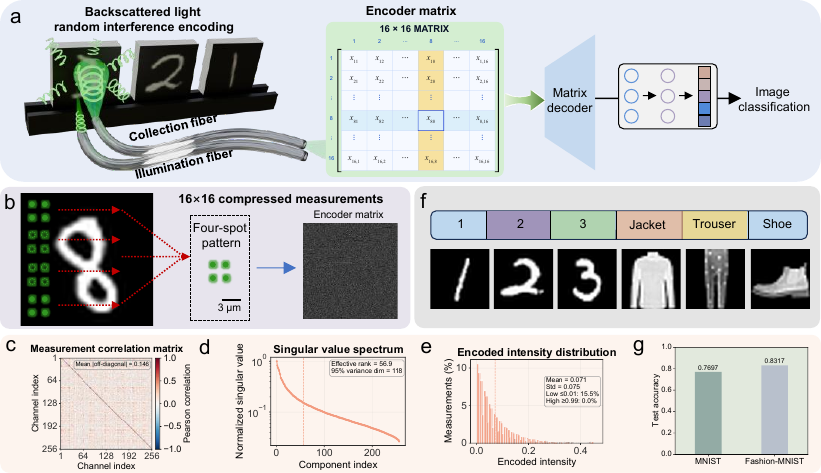}
\caption{Agent-proposed backscattered random-interference encoding. \textbf{a}, AHOIS proposes that object-dependent backscattered fields can be transformed by coherent scattering, multimode collection and intensity detection into a compact physical encoding. A four-focus illumination pattern is raster-scanned across the object, and the detected responses are assembled into a $16\times16$ encoder matrix for direct classification. \textbf{b}, Four-focus scanning and representative encoded measurement. \textbf{c}, Correlation structure of the encoded measurement ensemble. \textbf{d}, Singular-value spectrum and effective dimensionality of the physical encoding. \textbf{e}, Distribution of encoded intensities. \textbf{f}, Representative object classes used for neural decoding. \textbf{g}, Classification performance on MNIST and Fashion-MNIST.}
\label{fig2}
\end{figure*}

AHOIS then translated the hypothesis into an optical protocol. Gewu controlled the DMD and calibrated illumination fibre to generate a four-focus pattern at the object plane (Fig.~\ref{fig2}a). As this pattern was raster-scanned across the target, the simultaneously illuminated regions produced object-dependent backscattered fields that were collected by a separate multimode fibre. Each scan position yielded one intensity measurement after multimode propagation and detection, and the complete scan was arranged as a $16\times16$ encoder matrix (Fig.~\ref{fig2}a,b). The resulting matrix was treated as a compact optical representation of the object, not as a conventional image. Measurements were acquired from handwritten-digit and clothing targets derived from MNIST and Fashion-MNIST.

The measured responses satisfied the principal predictions of the hypothesis. The correlation matrix showed that the encoder matrices were not mutually identical and did not reduce to a single common intensity pattern (Fig.~\ref{fig2}c). The singular-value spectrum contained a broad set of non-negligible components, with an effective rank of 56.9, indicating that the measurements occupied a multidimensional feature space rather than collapsing onto a small number of trivial intensity variables (Fig.~\ref{fig2}d). The encoded-intensity distribution further showed that the detector response was populated over a usable dynamic range, with no dominant saturation regime (Fig.~\ref{fig2}e). These observations supported the interpretation that structured illumination, coherent backscattering, multimode propagation and square-law detection jointly redistributed object information into a speckle-like measurement space.

Having established this physical interpretation, AHOIS proposed neural decoding as a task-level test of whether the encoded measurements retained object information. A compact classifier was trained directly on the $16\times16$ encoder matrices, without using a reconstructed image as an intermediate representation. The classifier achieved test accuracies of 76.97\% on MNIST and 83.17\% on Fashion-MNIST (Fig.~\ref{fig2}f,g). Although the encoded matrices bore little visual resemblance to the corresponding objects, they retained sufficient information for object-level classification. This result showed that the backscattered speckle response was not merely an uninterpretable nuisance, but a computationally accessible representation produced by the physical measurement process.

This experiment illustrates the role of Socratic interrogation in the discovery loop. The encoding concept was not supplied as a target task; it emerged from the agent's analysis of unused backscattered light in the existing optical platform. Duzhi did not provide the answer, but forced the proposed mechanism to be evaluated against physical and statistical failure modes: redundancy, low dimensionality and unusable intensity range. The resulting measurement principle reframed backscattered speckle from an unwanted by-product of multimode-fibre imaging into a task-oriented physical encoding resource.

\section{Autonomous discovery of adaptive sparse scanning}\label{sec4}

AHOIS next identified a measurement-efficiency problem that arises from the sequential nature of multimode-fibre image acquisition. In conventional point-scanning imaging, an $N\times N$ reconstruction is obtained by positioning a focused illumination spot at $N^2$ spatial locations. This uniform raster strategy assigns the same measurement value to every position, although most scenes contain spatially non-uniform information: structural edges, high-contrast regions, fluorescent emitters or task-relevant objects occupy only part of the field of view. Because each measurement interrogates a localized region of the target, Boxue reasoned that a sparse preliminary scan could first reveal the coarse distribution of informative content, after which additional focus positions could be allocated selectively to regions where further measurements were expected to improve recognition or reconstruction. AHOIS therefore proposed a coarse-to-fine acquisition policy in which the scan pattern is adapted during the experiment, rather than fixed before acquisition.

Duzhi then challenged whether this strategy would be distinguishable from ordinary uniform subsampling. In particular, it questioned whether a coarse scan could miss small, weakly contrasted or spatially isolated features, and whether local signal intensity alone was sufficient to define where additional measurements should be placed. This interrogation led Boxue to refine the policy into three steps. First, an ultra-sparse background scan estimates the global field of view and identifies candidate regions of interest (ROIs). Second, the ROI is sampled more densely than the surrounding background, so that measurement effort is concentrated where structural or task-level information is present. Third, local densification is stopped when additional scan points no longer provide meaningful improvement in the reconstruction or recognition metric~\cite{chen2022}. The resulting strategy differs from fixed sparse sampling because the measurement distribution is determined by the scene itself and updated from the evidence already acquired.

\begin{figure*}[h]
\centering
\includegraphics[width=0.95\textwidth]{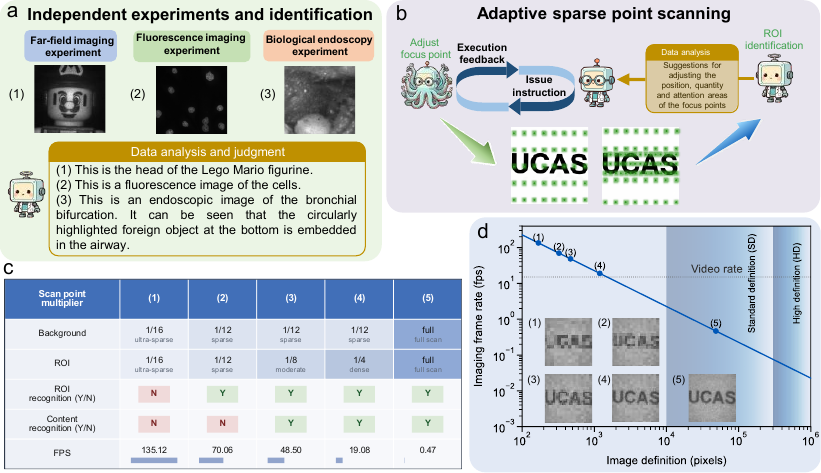}
\caption{Agent-proposed adaptive sparse point scanning. \textbf{a}, AHOIS analyses representative far-field, fluorescence and biological endoscopy measurements, identifying the corresponding scene content and task-relevant regions. \textbf{b}, Closed-loop sparse-scanning workflow. An initial low-density scan provides execution feedback, after which the agent identifies ROIs and adjusts the position, number and attention area of subsequent focus points. \textbf{c}, Sampling-density comparison across five acquisition settings. Ultra-sparse sampling of both background and ROI gives the highest frame rate but fails to support ROI or content recognition; progressively denser ROI sampling recovers recognition while retaining substantially higher speed than full scanning. \textbf{d}, Trade-off between imaging frame rate and effective image definition. Adaptive sparse acquisition maintains video-rate operation at reduced pixel counts, whereas full high-definition scanning provides the highest image definition at a much lower frame rate.}
\label{fig3}
\end{figure*}

AHOIS implemented this policy across multiple imaging conditions (Fig.~\ref{fig3}a,b). In the far-field imaging experiment, the system recognized the imaged target as the head of a Lego Mario figurine. In the fluorescence experiment, it identified localized cellular emission features. In the biological endoscopy experiment, it interpreted the scene as a bronchial bifurcation and highlighted a circular foreign object embedded near the lower airway region. These examples provided distinct scene structures for testing adaptive sampling: a reflective object with extended spatial features, a fluorescence image with sparse emitters and an endoscopic scene with localized clinically relevant content. Gewu translated the proposed sampling policy into focus-position sequences, Mingde monitored acquisition validity and Qiushi analysed the initial sparse measurements to estimate ROIs and recommend the next set of scan positions.

The sampling-density comparison in Fig.~\ref{fig3}c shows how recognition capability emerged as measurements were reallocated from the background to the ROI. In the most aggressive sparse condition, both the background and ROI were sampled at $1/16$ of the full scan density. This setting achieved the highest frame rate, 135.12 fps, but neither ROI recognition nor content recognition was successful. Increasing both background and ROI sampling to $1/12$ preserved a high frame rate of 70.06 fps and enabled ROI recognition, but the content itself was still not reliably recognized. When the background remained at $1/12$ density but the ROI was increased to $1/8$, both ROI and content recognition became successful while maintaining 48.50 fps. Further increasing the ROI density to $1/4$ improved the sampled representation but reduced the frame rate to 19.08 fps. Full scanning of both background and ROI provided complete spatial sampling but reduced the frame rate to 0.47 fps. Thus, the adaptive strategy identified an intermediate operating regime in which task-relevant recognition was retained at tens of frames per second, whereas full scanning was more than two orders of magnitude slower.

The frame-rate curve in Fig.~\ref{fig3}d further illustrates this trade-off. As the effective image definition increased from sparse representations to full high-definition sampling, the imaging frame rate decreased approximately monotonically. The sparse and ROI-adaptive settings occupied the video-rate region, whereas the full-scan setting moved into a low-frame-rate regime despite producing the highest pixel definition. The representative reconstructions show that the most aggressive sparse setting produced block-like, weakly recognizable structure, whereas moderate ROI densification recovered the principal ``UCAS'' content while avoiding the cost of full-field densification. These results indicate that, for scenes with localized informative structure, the critical variable is not the total number of sampled positions but whether the measurements are placed where they contribute to the task.

This experiment therefore reframed sparse acquisition as evidence-directed measurement allocation. The adaptive policy was not provided as a predefined algorithm; it arose from the agent's physical understanding that the MMF system acquires spatial information through localized optical interactions and that the value of an additional measurement depends on both its position and the information already collected. Socratic questioning strengthened this proposal by exposing the failure modes of naive subsampling, especially missed small features and intensity-only ROI selection. The resulting strategy allowed AHOIS to choose where to observe next from the evolving content of the experiment, converting scan-density reduction from a passive compression step into an active decision process.

\section{Autonomous diagnosis and paper-to-platform reproduction with AHOIS}\label{sec5}

\begin{figure*}[h]
\centering
\includegraphics[width=0.95\textwidth]{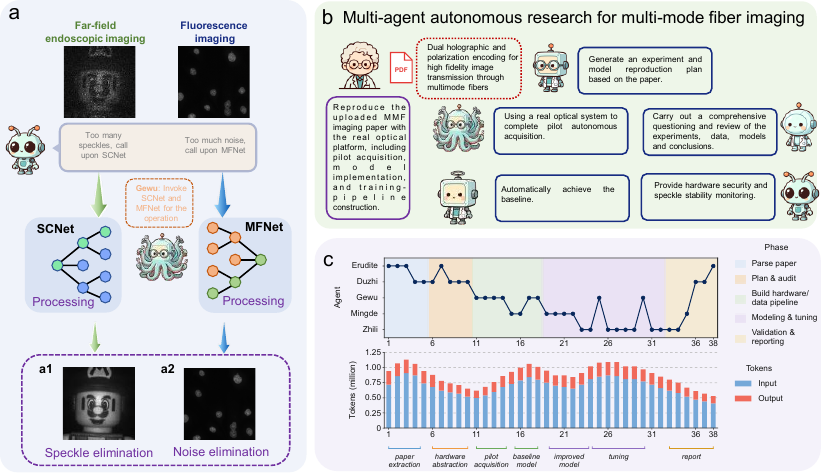}
\caption{Autonomous diagnosis and paper-to-platform reproduction. \textbf{a}, AHOIS distinguishes speckle-dominated degradation in coherent far-field imaging from noise-dominated degradation in fluorescence imaging and invokes SCNet and MFNet as modality-specific processing tools. \textbf{b}, The multi-agent system translates a published multimode-fibre reconstruction study into a local workflow comprising paper interpretation, experimental planning, pilot optical acquisition, reconstruction-model implementation, training and evaluation. Large-scale repetitive acquisition was subsequently performed by the researchers using the agent-established procedure. \textbf{c}, Computational trace of one complete agent-led reproduction trajectory. The upper panel records the active agent across 38 steps, and the lower panel shows input and output token consumption across paper parsing, hardware abstraction, pilot acquisition, baseline modelling, model improvement, tuning and reporting.}
\label{fig4}
\end{figure*}

We next examined whether AHOIS could use its acquired physical model of the multimode-fibre platform to diagnose experimental failure modes and translate external scientific knowledge into an executable local workflow. The two imaging modalities produced qualitatively different forms of degradation. In coherent far-field imaging, the dominant degradation appeared as spatially distributed speckle that obscured the underlying image structure. In fluorescence imaging, the measured frames were instead dominated by weak signals and detector-related noise. AHOIS did not apply a generic enhancement operation to both cases. It associated each degradation pattern with its corresponding physical modality and selected a modality-specific computational intervention: SCNet\cite{zeng2026physics} for speckle suppression in far-field images and fluorescence image denoising network (MFNet) for denoising fluorescence measurements (Fig.~\ref{fig4}a). The processed images were then returned to the analysis loop for reassessment. This experiment tested whether the agents could move from observation to physical diagnosis and tool selection, rather than merely executing a predetermined image-processing step.

We then tested whether AHOIS could convert a published multimode-fibre imaging protocol~\cite{Xu2025Dual} into an operational procedure on the local platform. The selected study used holographic and polarization encoding on a digital micromirror device, transmitted the encoded fields through a multimode fibre, recorded the resulting speckle measurements and reconstructed the input handwritten-digit images using a U-Net-like inverse model~\cite{ronneberger2015}. The task required more than reproducing a neural-network architecture. AHOIS had to identify which elements of the reported method represented the essential measurement principle, determine which elements were implementation-specific, and adapt the optical encoding, hardware control, acquisition sequence and reconstruction pipeline to a different experimental configuration.

Starting from the source paper and a description of the available instrument, Boxue reconstructed the experimental logic linking input-image encoding, DMD pattern generation, multimode-fibre propagation, camera acquisition and computational reconstruction. Duzhi then questioned assumptions that could not be transferred directly from the published apparatus to the local system, including differences in hardware access, optical alignment, calibration state and data-acquisition scale. This interrogation prompted Boxue to separate the general measurement principle from apparatus-specific details. Gewu translated the resulting plan into executable routines for optical-pattern generation, DMD control and camera acquisition. Mingde monitored the operating state of the platform during pilot execution, including acquisition completeness and speckle stability. Qiushi prepared the data-processing, model-implementation and training components required to complete the reconstruction workflow (Fig.~\ref{fig4}b).

The autonomous optical acquisition was deliberately performed at pilot scale. This pilot experiment verified that the agents could execute the full pathway from encoded input generation to physical acquisition and data transfer into the reconstruction pipeline. We did not require AHOIS to repeat the validated acquisition sequence for the entire training set, because large-scale data collection mainly consisted of time-consuming repetition and did not introduce a new scientific-reasoning challenge. The remaining measurements were therefore collected by the researchers using the agent-established procedure. Human participation replaced the repetitive scaling of acquisition, whereas the interpretation of the paper, adaptation of the protocol, pilot execution and construction of the computational workflow remained directed by AHOIS.

The reconstruction model was not manually specified for AHOIS. After analysing the source paper and the acquired MMF measurements, Qiushi first implemented a U-Net-like baseline following the reported reconstruction principle, with an encoder--decoder structure, skip connections, mean absolute error loss, Adam optimization and a sigmoid-constrained grayscale output. It then proposed an improved residual-attention architecture to recover structural information more effectively from the highly transformed speckle measurements. The modified network incorporated residual convolutional blocks, channel-attention modules and a compound loss combining pixel-level reconstruction error with structural similarity. The baseline and improved models were evaluated using the same train--validation--test partition and the same quantitative metrics, including SSIM, PSNR and pixel-wise correlation.

Qiushi generated the principal model and training code, selected the initial optimizer, learning rate, batch size, training duration and loss composition, and adjusted the training strategy when intermediate reconstructions or optimization behaviour were unsatisfactory. The researchers supplied the scaled measurements and maintained the computing environment, but the architecture selection, implementation strategy and principal training decisions were produced by the agent. This result showed that AHOIS could progress from understanding the physical objective of a source study to constructing and optimizing an inverse model for measurements acquired on the local optical platform.

The computational trace in Fig.~\ref{fig4}c provides an audit of how responsibility shifted across the agents during one end-to-end paper-to-platform trajectory. The process comprised 38 recorded steps spanning paper extraction, plan auditing, hardware abstraction, pilot acquisition, baseline-model construction, improved-model development, tuning and reporting. Early steps were dominated by Boxue and Duzhi, reflecting the need to reconstruct the scientific logic of the paper, identify transferable assumptions and adapt the protocol. Gewu and Mingde became active when the plan was converted into hardware operations and tested on the real instrument. Qiushi dominated the later stages associated with reconstruction-model implementation, training supervision and evaluation. The token bars show that input context accounted for most of the computational load at each step, whereas output generation was smaller and more variable, consistent with a workflow in which agents repeatedly consumed paper content, system state and experimental feedback before producing targeted plans or code.

This experiment distinguishes autonomous scientific reasoning from exhaustive autonomous labour. AHOIS was not asked to spend resources repeating an already validated acquisition sequence. Instead, it demonstrated that it could interpret a published optical method, identify its transferable measurement principle, adapt it to a non-original platform, validate the physical acquisition path at pilot scale, construct the reconstruction pipeline and direct most of the model-optimization process. In combination with the modality-specific diagnosis in Fig.~\ref{fig4}a, this result shows that the agents could move from observing a problem to identifying its physical character, selecting an appropriate intervention and converting published scientific knowledge into an operational procedure on real hardware.

\section{Socratic midwifery improves physical understanding}\label{sec6}

We next evaluated whether the Socratic component of AHOIS improved the quality of physical reasoning, rather than merely increasing response length or adding another review step. Scientific reasoning in AHOIS was not implemented as a single-pass generation process~\cite{wei2022,yao2023,shinn2023}. Boxue's interpretations were instead iteratively examined by Duzhi under scientist supervision. Duzhi was constrained to question rather than propose candidate mechanisms, research directions or preferred conclusions. Its role was to expose ambiguous definitions, unsupported assumptions, unresolved contradictions and incomplete causal links in Boxue's current account of the optical system. This design allowed us to test whether Boxue could revise its own physical explanation in response to structured interrogation, rather than reproduce an answer supplied by the critic.

\begin{figure*}[h]
\centering
\includegraphics[width=0.95\textwidth]{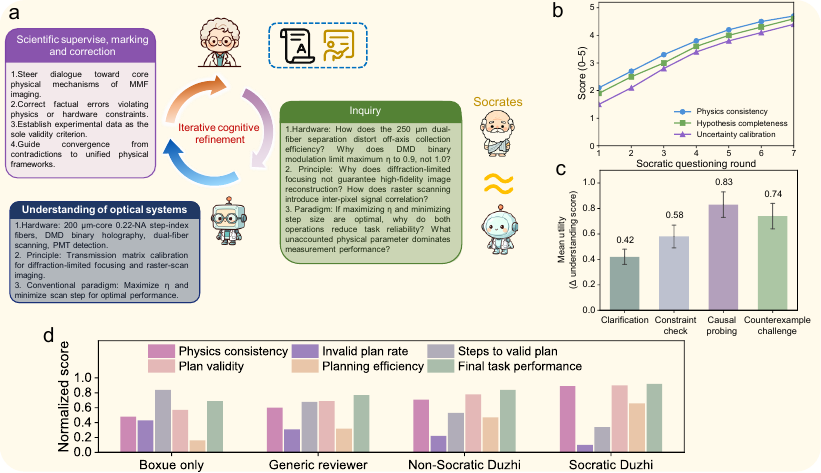}
\caption{Socratic midwifery improves physics-grounded reasoning. \textbf{a}, Duzhi examines Boxue's interpretation through clarification, physical-constraint checking, causal probing and counterexample-based questioning without supplying candidate scientific conclusions. \textbf{b}, Physics consistency, hypothesis completeness and uncertainty calibration improve over successive Socratic questioning rounds. \textbf{c}, Mean improvement in the physical-understanding score associated with the four classes of Socratic intervention. Causal probing and counterexample challenge produce the largest gains. \textbf{d}, Ablation across Boxue alone, a generic reviewer, non-Socratic Duzhi and Socratic Duzhi, evaluated using normalized physics consistency, plan validity, planning efficiency and downstream task performance. Invalid-plan rate and steps to valid plan are inverted during normalization so that higher values indicate better performance.}
\label{fig5}
\end{figure*}

The questioning protocol operated at four levels (Fig.~\ref{fig5}a). Clarification questions required Boxue to define the physical meaning of terms in its explanation and to distinguish measured quantities from inferred or assumed quantities. Constraint-checking questions examined whether the proposed interpretation was compatible with optical principles and hardware limits, including energy transfer, coherence, multimode propagation, collection geometry and detector response. Causal-probing questions required Boxue to describe the full chain linking optical modulation, field propagation, object interaction, signal collection and final measurement. Counterexample-based questions tested whether the same explanation would remain valid when the object, imaging modality, sampling condition or system stability changed. These interventions targeted the structure of the reasoning rather than the content of a particular hypothesis, and therefore did not prescribe the scientific conclusion.

We quantified the effect of successive questioning rounds using three scores: physics consistency, hypothesis completeness and uncertainty calibration. Physics consistency assessed whether the explanation followed relevant optical principles without internal contradiction. Hypothesis completeness assessed whether the explanation connected the main components of the experiment, including illumination, propagation, object interaction, collection and detection, rather than attributing the observation to a single isolated factor. Uncertainty calibration assessed whether Boxue separated experimentally established facts from model-dependent interpretations and unresolved propositions requiring further evidence. All three scores increased over seven rounds of Socratic questioning (Fig.~\ref{fig5}b). Physics consistency rose from approximately 2 to above 4.5 on the 0--5 scale, while hypothesis completeness and uncertainty calibration showed similar monotonic improvements. These trends indicate that the questioning process progressively converted an incomplete qualitative account into a more constrained and testable physical explanation.

The four classes of Socratic intervention contributed unequally (Fig.~\ref{fig5}c). Clarification produced a mean utility of 0.42 by removing ambiguous terminology and forcing Boxue to state what was measured, inferred or assumed. Constraint checking produced a larger gain of 0.58 by excluding explanations that conflicted with optical principles or hardware constraints. Causal probing produced the largest improvement, 0.83, because it forced the agent to connect individual observations to a complete physical pathway rather than rely on qualitative association. Counterexample challenge also produced a strong gain of 0.74 by testing whether the proposed explanation remained valid outside the conditions in which it was first formulated. The improvement therefore did not arise simply from adding more criticism. It arose from questions that changed the internal structure of the hypothesis by making causal dependencies, applicability conditions and failure modes explicit.

A representative dialogue concerned the assumption that a sharper optical focus and a denser raster scan should necessarily produce a better measurement. Boxue initially treated focal concentration and spatial sampling density as direct indicators of imaging performance. Duzhi did not provide an alternative conclusion. Instead, it asked whether this interpretation accounted for collection efficiency, object-dependent scattering, detector response and redundancy between neighbouring measurements. It further asked whether increasing the number of scan positions would continue to add independent information when adjacent measurements overlapped, and whether an apparently optimal illumination condition could still lead to poorer task-level performance because of limitations elsewhere in the measurement chain. Boxue consequently revised its explanation from a single-factor account to a system-level model in which illumination, object interaction, collection, detection and information redundancy jointly determine measurement utility. This revised account supported more defensible experimental choices without the critic prescribing those choices directly.

To isolate the contribution of Socratic structure, we compared four configurations: Boxue alone, Boxue paired with a generic reviewer, Boxue paired with a non-Socratic version of Duzhi and Boxue paired with the full Socratic Duzhi. The Socratic configuration achieved the highest normalized physics consistency, plan validity, planning efficiency and final task performance (Fig.~\ref{fig5}d). It also produced the lowest invalid-plan rate and required fewer steps to reach a valid plan; these two metrics were inverted during normalization so that higher values consistently represented better outcomes. Generic review improved surface-level completeness but was less effective at exposing missing causal links or specifying conditions under which a conclusion could fail. Non-Socratic Duzhi identified weaknesses more directly, but often replaced Boxue's reasoning with corrective statements. In contrast, Socratic Duzhi forced Boxue to resolve inconsistencies itself, preserving the distinction between guided inquiry and answer provision.

These results show that Duzhi's contribution lies not in the amount of criticism but in the form of the criticism. By withholding direct solutions and focusing on definitions, physical constraints, causal structure and counterexamples, Duzhi encouraged Boxue to construct explanations that were more internally consistent, more complete and more closely connected to experimental decisions. Socratic midwifery therefore provides a mechanism for improving the epistemic quality of an autonomous agent without predetermining the hypothesis that the agent is expected to reach~\cite{cui2026inquitree,li2026causalagent,ho2026soundness}.

\section{Discussion}\label{sec7}

We have demonstrated that a multi-agent AI system can carry out closed-loop experimental reasoning on a real high-dimensional optical platform. AHOIS did not merely execute predefined measurement scripts. It formulated physical hypotheses about the instrument, converted them into testable experimental procedures, diagnosed modality-dependent failure modes, adapted a published multimode-fibre imaging protocol to a non-original configuration and improved the consistency of its physical explanations through structured Socratic interrogation. These results show that agentic autonomy in experimental science can extend beyond workflow coordination towards the construction, criticism and revision of evidence-grounded physical accounts.

The central advance of AHOIS is the introduction of Socratic midwifery as an operational mechanism for epistemic autonomy. Existing agentic systems increasingly support literature synthesis, hypothesis generation, protocol design, instrument control and data analysis. However, in many cases the scientific frame is still supplied by human researchers: the objective, action space, evaluation metric and interpretation route are defined before the agent begins. AHOIS addresses a different problem. It asks whether an artificial system can examine ambiguous experimental evidence, expose hidden assumptions, distinguish competing physical explanations and decide which intervention would make those explanations testable. The role of Duzhi is therefore not to add another reviewer to the workflow, but to transform candidate explanations into explicit, constrained and falsifiable scientific states.

The multimode-fibre platform was selected not as the final application domain, but as a representative complex physical system. It combines high-dimensional optical control, dense mode mixing, indirect intensity measurements, calibration sensitivity, environmental drift and modality-dependent noise. These features make interpretation non-trivial: a change in the measured signal may arise from useful physical encoding, uncontrolled decorrelation, detector noise, fluorescence contamination, imperfect alignment or a failure in execution. AHOIS used this ambiguity as a testbed for scientific reasoning. Its discovery of backscattered interference encoding, its derivation of adaptive sparse scanning and its modality-specific selection of image-processing tools show how physical understanding can guide measurement design and experimental decision-making in systems where the measured response is not directly interpretable.

The framework should be transferable to other experimental settings that share these properties. Adaptive microscopes, scattering-based imaging systems, photonic processors, autonomous synthesis platforms and high-throughput materials-characterization systems all require decisions under partial observability, instrument drift and competing mechanistic interpretations~\cite{raj2024,boiko2023,szymanski2023,zhang2026agentic,song2026catalyst,jiang2026deltaevolve}. In such environments, the limiting factor is often not whether an agent can issue a command to an instrument, but whether it can determine what the resulting measurement means. The separation of scientific planning, hardware abstraction, integrity monitoring, modelling and Socratic criticism provides a general architectural principle for this class of problems.

Several limitations remain. First, Socratic interrogation is currently implemented in natural language. This provides flexibility, but it also introduces ambiguity in the definition of assumptions, constraints and falsification criteria. Future systems could ground the questioning process in formal representations of physical variables, causal graphs, optical models or machine-verifiable experimental protocols. Second, human supervision remains part of the system. In the present work, the supervisor acted as a safety and quality gate rather than as the source of the scientific hypotheses, but stronger autonomy will require formal verification of hardware actions, explicit risk models and auditable intervention policies. Third, AHOIS does not yet claim to invent new physical laws. Its current strength is to construct, discriminate and revise mechanistic explanations within a known physical domain. Extending this approach to open-ended hypothesis generation beyond a predefined conceptual library remains an important challenge.

The present experiments also expose a practical distinction between autonomous reasoning and exhaustive autonomous labour. In the paper-to-platform experiment, AHOIS performed the intellectual and procedural translation of a published protocol and validated the acquisition path at pilot scale, while researchers carried out the repetitive large-scale acquisition using the agent-established procedure. This boundary is deliberate. The scientific question was whether the agent could understand, adapt and operationalize the method, not whether it should consume resources repeating an already validated acquisition sequence. Future work should quantify this boundary more systematically by separating reasoning-intensive steps from scalable robotic execution.

More broadly, our results suggest that trustworthy AI for experimental science requires more than fluent hypothesis generation or reliable instrument control. A scientific agent must be able to state what it assumes, identify what it has measured, distinguish evidence from interpretation, recognize when an explanation has failed and choose the next experiment accordingly. AHOIS provides one route towards this goal by combining separated agent roles, real hardware feedback, system-integrity monitoring and Socratic criticism. Such systems will not replace the scientist as a source of judgment in the near term, but they may become increasingly capable collaborators in experiments where the central challenge is to decide what a complex instrument is revealing~\cite{krenn2022,wang2024agent,sumers2024,lin2026compphysics,somasekharan2026aicfd}.

\section{Methods}\label{sec8}

\subsection{Multimode-fibre optical platform}

The experimental platform combined far-field multimode-fibre imaging and fluorescence imaging within a common agent-controlled optical framework. For far-field imaging, a linearly polarized laser was modulated by a digital micromirror device (DMD; $1024\times768$ pixels) using binary Lee holograms. The active DMD region was restricted to $624\times624$ pixels. The modulated field was relayed to a step-index illumination multimode fibre (MMF) with a core diameter of 100~$\mu$m and a numerical aperture of 0.22. Polarization-diverse modulation was used to increase the accessible input-mode space. At the distal side, the MMF output was projected into the far field to generate programmable focal spots at selected spatial positions. Backscattered light from the object was collected by an adjacent MMF and routed to the detection path.

The fluorescence configuration used a related fibre-scanning geometry. Excitation was delivered to the sample through the optical system, and emitted photons were separated spectrally before detection. The two modalities therefore shared the same general principle of spatially controlled excitation and sequential acquisition, but their signals were governed by different physical processes. Far-field reflectance imaging was dominated by coherent backscattering, speckle formation and multimode interference, whereas fluorescence imaging was governed by incoherent, photon-limited emission and detector noise. This distinction was used by AHOIS when diagnosing modality-dependent degradation and selecting computational processing tools.

\subsection{Transmission-matrix calibration and focus generation}

The illumination MMF was calibrated by measuring its complex transmission matrix using phase-stepped interferometry with an external reference beam. Binary amplitude gratings displayed on the DMD generated a set of input plane-wave modes, and the corresponding distal fields were recorded on a camera. Approximately 7,500 input patterns were used to construct the calibrated input basis. The output field was sampled on an oversampled camera grid containing approximately 173,000 addressable positions.

The measured transmission matrix was used to calculate DMD holograms for programmable far-field focusing. Transmission-matrix acquisition, reconstruction, hologram synthesis and transfer of the resulting patterns to the DMD were completed within several minutes. Only the illumination fibre was calibrated for focus generation. The adjacent MMF served as the collection channel and was not used for generating the illumination focus.

\subsection{AHOIS implementation and agent--instrument interface}

AHOIS interacted with the optical platform through instrument-control functions for hologram generation, DMD pattern loading, detector acquisition, metadata recording and experimental-parameter adjustment. Boxue generated scientific hypotheses and machine-readable experimental plans. Gewu translated these plans into executable optical operations, including DMD control, pattern sequencing and acquisition triggering. Mingde monitored basic acquisition validity, including detector saturation, exposure range, dark-frame validity, frame completeness and speckle repeatability. Qiushi performed data loading, model implementation, reconstruction, training supervision and quantitative analysis. Duzhi acted as a Socratic physics critic, interrogating Boxue's assumptions, causal links, physical constraints and possible failure modes.

The agents exchanged information through a shared experimental state that recorded the current hypothesis, assumptions, proposed action, expected observation, acquired data, uncertainty and next decision. Short pilot acquisitions were performed before extended experiments to confirm that the proposed optical and computational pathway could be executed on the physical system. Agent states, experimental plans, tool calls and acquired results were logged for later inspection and replay. The underlying language-model configuration, tool permissions, prompt templates and representative agent trajectories are reported in the Supplementary Information.

\subsection{Backscattered interference encoding}

For backscattered interference encoding, the calibrated illumination MMF generated a four-focus pattern at the object plane. The pattern was raster-scanned across each target. At each scan position, the simultaneously illuminated regions generated object-dependent backscattered fields, which were collected by the adjacent MMF. Propagation and modal mixing within the collection fibre transformed the backscattered field before square-law detection. Each scan position therefore produced one encoded intensity value. The complete scan was arranged into a $16\times16$ encoder matrix.

The encoder matrices were not intended to reproduce visually recognizable images. Instead, they were treated as compact object-dependent optical measurements generated by structured illumination, coherent backscattering, multimode propagation and intensity detection. Experiments were conducted using handwritten-digit and clothing targets derived from MNIST and Fashion-MNIST.

\subsection{Analysis of encoded measurements}

The statistical structure of the encoded measurements was assessed using correlation analysis, singular-value decomposition and intensity-distribution analysis. Pairwise correlation coefficients were assembled into a correlation matrix to evaluate whether different inputs produced distinct and non-redundant encoded responses. Singular-value decomposition was applied to the measurement ensemble to characterize how information was distributed across independent components. The effective rank was calculated from the normalized singular-value spectrum. Encoded-intensity distributions were analysed to determine whether the detector range was used effectively and whether the measurements were dominated by near-zero values, saturation or uniform background.

A compact neural classifier was trained directly on the $16\times16$ encoder matrices to test whether the optical encoding retained object-level information. No reconstructed image was used as an intermediate representation. Classification performance was evaluated on held-out test data. Reported classification accuracies are the mean over five random initializations, and confidence intervals are provided in the Supplementary Information.

\subsection{Adaptive sparse scanning}

For adaptive sparse scanning, AHOIS implemented a coarse-to-fine acquisition strategy derived from the localized interaction between the scanning focus and the object. An initial low-density scan, typically using an $8\times8$ grid, provided a coarse representation of the field of view. Qiushi analysed this preliminary representation to estimate candidate regions of interest (ROIs). Gewu then generated a second scan with increased sampling density within the selected ROIs while keeping the background more sparsely sampled.

Local densification continued until the incremental improvement in the reconstruction quality or task-related metric fell below a predefined threshold. The adaptive protocol was compared with full raster scanning and fixed uniform subsampling under consistent imaging conditions. For far-field imaging, informative regions were identified from spatial structure and reflectance contrast. For fluorescence imaging, informative regions were identified from localized emission features. Frame rate, sampled-pixel count, ROI recognition and content recognition were used to compare acquisition settings.

\subsection{Autonomous image diagnosis and modality-specific processing}

During autonomous image diagnosis, AHOIS evaluated the observed degradation in relation to the underlying imaging modality. Speckle-dominated far-field images were processed using SCNet, whereas noise-dominated fluorescence images were processed using MFNet. The processing model was selected by the agents from the measured image characteristics and the inferred physical origin of the degradation, rather than prescribed manually for each experiment. Processed images were returned to the analysis loop for interpretation and task-level evaluation.

\subsection{Paper-to-platform reproduction}

The paper-to-platform experiment used a published MMF reconstruction study as the source protocol. AHOIS extracted the relation between holographic and polarization encoding, DMD modulation, MMF transmission, speckle acquisition and computational reconstruction. The agents then adapted this sequence to the available local platform.

The agent-led trajectory included paper interpretation, experimental planning, protocol adaptation, pattern generation, pilot optical acquisition, data transfer, reconstruction-model implementation, training and performance evaluation. The autonomous optical acquisition was deliberately performed at pilot scale to verify that the complete physical and computational pathway could be executed. Because full acquisition of the training set required repetition of the validated procedure rather than additional scientific reasoning, the remaining large-scale measurements were collected by the researchers using the agent-established acquisition method.

\subsection{Reconstruction models and training}

Qiushi first implemented a U-Net-like reconstruction baseline with an encoder--decoder architecture, skip connections, mean absolute error loss, Adam optimization and a sigmoid-constrained grayscale output~\cite{ronneberger2015}. It then proposed a residual-attention architecture designed to recover structural information more effectively from highly transformed MMF speckle measurements. The modified model incorporated residual convolutional blocks, channel-attention modules and a compound loss combining pixel-level reconstruction error with structural similarity~\cite{he2016,vaswani2017}.

Qiushi generated the principal model and training code, selected the initial optimizer, learning rate, batch size, loss composition and training duration, and adjusted the main training parameters during optimization. The researchers supplied the scaled dataset and maintained the computing environment. Network design, implementation strategy and principal optimization decisions were produced by AHOIS. The baseline and improved models were evaluated using the same train--validation--test partition. Reconstruction performance was quantified using structural similarity index measure (SSIM), peak signal-to-noise ratio (PSNR) and pixel-wise correlation on held-out test data.

\subsection{Evaluation of Socratic midwifery}

The effect of Socratic midwifery was quantified using physics consistency, hypothesis completeness and uncertainty calibration. Physics consistency assessed whether a generated explanation obeyed the relevant optical principles and remained internally coherent. Hypothesis completeness assessed whether the explanation connected illumination, propagation, object interaction, collection and detection into a sufficiently complete physical account. Uncertainty calibration assessed whether experimentally established observations were distinguished from inferred mechanisms and unresolved propositions.

Responses were evaluated on a five-point scale according to a predefined rubric. Four configurations were compared using identical scientific prompts: Boxue alone, Boxue with a generic reviewer, Boxue with a non-Socratic critic and Boxue with Socratic Duzhi. Additional evaluation metrics included experimental-plan validity, invalid-plan rate, number of reasoning steps required to obtain an acceptable plan, planning efficiency and downstream task performance. Invalid-plan rate and steps to a valid plan were inverted during normalization so that higher normalized values consistently represented better performance.

\subsection{Token accounting}

Token consumption was measured across the five agents during one complete paper-to-platform trajectory. The reported token distribution included paper interpretation, scientific planning, protocol adaptation, pilot acquisition, model construction, code generation, training supervision and performance evaluation. It excluded the subsequent large-scale repetitive acquisition performed by the researchers. Token counts were aggregated by agent and by workflow phase to characterize how computational effort shifted from scientific interpretation to hardware execution and model development.

\subsection{Statistical analysis}

Classification results are reported as the mean over five random initializations. Reconstruction metrics were calculated on held-out test data using a fixed dataset partition shared by all compared models. Confidence intervals, sample sizes and additional statistical details are provided in the Supplementary Information. Sample sizes were selected to provide representative coverage of the imaging and classification tasks and were not determined using a prospective statistical-power calculation.

\backmatter

\section{Acknowledgements}\label{sec9}
X.R.Z. and Y.D. acknowledge support from the Hangzhou Joint Fund of the Zhejiang Provincial Natural Science Foundation of China (Grant No. LHZY24F030002), Funds of the Natural Science Foundation of Hangzhou (Grant No. 2025SZRJJ1384), Open Fund of the State Key Laboratory of Precision Meauring Technology and Instruments(Tianjin University, Grant No. Pilab2401), and funding from the Hangzhou Institute for Advanced Study, UCAS.

\section{Competing interests}\label{sec10}
The other authors declare no competing interests.
\bigskip


\bibliography{sn-bibliography}


\begin{thebibliography}{63}
\ifx \bisbn   \undefined \def \bisbn  #1{ISBN #1}\fi
\ifx \binits  \undefined \def \binits#1{#1}\fi
\ifx \bauthor  \undefined \def \bauthor#1{#1}\fi
\ifx \batitle  \undefined \def \batitle#1{#1}\fi
\ifx \bjtitle  \undefined \def \bjtitle#1{#1}\fi
\ifx \bvolume  \undefined \def \bvolume#1{\textbf{#1}}\fi
\ifx \byear  \undefined \def \byear#1{#1}\fi
\ifx \bissue  \undefined \def \bissue#1{#1}\fi
\ifx \bfpage  \undefined \def \bfpage#1{#1}\fi
\ifx \blpage  \undefined \def \blpage #1{#1}\fi
\ifx \burl  \undefined \def \burl#1{\textsf{#1}}\fi
\ifx \doiurl  \undefined \def \doiurl#1{\url{https://doi.org/#1}}\fi
\ifx \betal  \undefined \def \betal{\textit{et al.}}\fi
\ifx \binstitute  \undefined \def \binstitute#1{#1}\fi
\ifx \binstitutionaled  \undefined \def \binstitutionaled#1{#1}\fi
\ifx \bctitle  \undefined \def \bctitle#1{#1}\fi
\ifx \beditor  \undefined \def \beditor#1{#1}\fi
\ifx \bpublisher  \undefined \def \bpublisher#1{#1}\fi
\ifx \bbtitle  \undefined \def \bbtitle#1{#1}\fi
\ifx \bedition  \undefined \def \bedition#1{#1}\fi
\ifx \bseriesno  \undefined \def \bseriesno#1{#1}\fi
\ifx \blocation  \undefined \def \blocation#1{#1}\fi
\ifx \bsertitle  \undefined \def \bsertitle#1{#1}\fi
\ifx \bsnm \undefined \def \bsnm#1{#1}\fi
\ifx \bsuffix \undefined \def \bsuffix#1{#1}\fi
\ifx \bparticle \undefined \def \bparticle#1{#1}\fi
\ifx \barticle \undefined \def \barticle#1{#1}\fi
\bibcommenthead
\ifx \bconfdate \undefined \def \bconfdate #1{#1}\fi
\ifx \botherref \undefined \def \botherref #1{#1}\fi
\ifx \url \undefined \def \url#1{\textsf{#1}}\fi
\ifx \bchapter \undefined \def \bchapter#1{#1}\fi
\ifx \bbook \undefined \def \bbook#1{#1}\fi
\ifx \bcomment \undefined \def \bcomment#1{#1}\fi
\ifx \oauthor \undefined \def \oauthor#1{#1}\fi
\ifx \citeauthoryear \undefined \def \citeauthoryear#1{#1}\fi
\ifx \endbibitem  \undefined \def \endbibitem {}\fi
\ifx \bconflocation  \undefined \def \bconflocation#1{#1}\fi
\ifx \arxivurl  \undefined \def \arxivurl#1{\textsf{#1}}\fi
\csname PreBibitemsHook\endcsname

\bibitem[\protect\citeauthoryear{King et~al.}{2004}]{king2004functional}
\begin{barticle}
\bauthor{\bsnm{King}, \binits{R.D.}},
\bauthor{\bsnm{Whelan}, \binits{K.E.}},
\bauthor{\bsnm{Jones}, \binits{F.M.}},
\bauthor{\bsnm{Reiser}, \binits{P.G.}},
\bauthor{\bsnm{Bryant}, \binits{C.H.}},
\bauthor{\bsnm{Muggleton}, \binits{S.H.}},
\bauthor{\bsnm{Kell}, \binits{D.B.}},
\bauthor{\bsnm{Oliver}, \binits{S.G.}}:
\batitle{Functional genomic hypothesis generation and experimentation by a
  robot scientist}.
\bjtitle{Nature}
\bvolume{427}(\bissue{6971}),
\bfpage{247}--\blpage{252}
(\byear{2004})
\end{barticle}
\endbibitem

\bibitem[\protect\citeauthoryear{King et~al.}{2009}]{king2009automation}
\begin{barticle}
\bauthor{\bsnm{King}, \binits{R.D.}},
\bauthor{\bsnm{Rowland}, \binits{J.}},
\bauthor{\bsnm{Oliver}, \binits{S.G.}},
\bauthor{\bsnm{Young}, \binits{M.}},
\bauthor{\bsnm{Aubrey}, \binits{W.}},
\bauthor{\bsnm{Byrne}, \binits{E.}},
\bauthor{\bsnm{Liakata}, \binits{M.}},
\bauthor{\bsnm{Markham}, \binits{M.}},
\bauthor{\bsnm{Pir}, \binits{P.}},
\bauthor{\bsnm{Soldatova}, \binits{L.N.}}, \betal:
\batitle{The automation of science}.
\bjtitle{Science}
\bvolume{324}(\bissue{5923}),
\bfpage{85}--\blpage{89}
(\byear{2009})
\end{barticle}
\endbibitem

\bibitem[\protect\citeauthoryear{Schmidt and
  Lipson}{2009}]{schmidt2009distilling}
\begin{barticle}
\bauthor{\bsnm{Schmidt}, \binits{M.}},
\bauthor{\bsnm{Lipson}, \binits{H.}}:
\batitle{Distilling free-form natural laws from experimental data}.
\bjtitle{science}
\bvolume{324}(\bissue{5923}),
\bfpage{81}--\blpage{85}
(\byear{2009})
\end{barticle}
\endbibitem

\bibitem[\protect\citeauthoryear{Burger et~al.}{2020}]{burger2020mobile}
\begin{barticle}
\bauthor{\bsnm{Burger}, \binits{B.}},
\bauthor{\bsnm{Maffettone}, \binits{P.M.}},
\bauthor{\bsnm{Gusev}, \binits{V.V.}},
\bauthor{\bsnm{Aitchison}, \binits{C.M.}},
\bauthor{\bsnm{Bai}, \binits{Y.}},
\bauthor{\bsnm{Wang}, \binits{X.}},
\bauthor{\bsnm{Li}, \binits{X.}},
\bauthor{\bsnm{Alston}, \binits{B.M.}},
\bauthor{\bsnm{Li}, \binits{B.}},
\bauthor{\bsnm{Clowes}, \binits{R.}}, \betal:
\batitle{A mobile robotic chemist}.
\bjtitle{Nature}
\bvolume{583}(\bissue{7815}),
\bfpage{237}--\blpage{241}
(\byear{2020})
\end{barticle}
\endbibitem

\bibitem[\protect\citeauthoryear{Szymanski
  et~al.}{2023}]{szymanski2023autonomous}
\begin{barticle}
\bauthor{\bsnm{Szymanski}, \binits{N.J.}},
\bauthor{\bsnm{Rendy}, \binits{B.}},
\bauthor{\bsnm{Fei}, \binits{Y.}},
\bauthor{\bsnm{Kumar}, \binits{R.E.}},
\bauthor{\bsnm{He}, \binits{T.}},
\bauthor{\bsnm{Milsted}, \binits{D.}},
\bauthor{\bsnm{McDermott}, \binits{M.J.}},
\bauthor{\bsnm{Gallant}, \binits{M.}},
\bauthor{\bsnm{Cubuk}, \binits{E.D.}},
\bauthor{\bsnm{Merchant}, \binits{A.}}, \betal:
\batitle{An autonomous laboratory for the accelerated synthesis of inorganic
  materials}.
\bjtitle{Nature}
\bvolume{624}(\bissue{7990}),
\bfpage{86}
(\byear{2023})
\end{barticle}
\endbibitem

\bibitem[\protect\citeauthoryear{Gottweis
  et~al.}{2026}]{gottweis2026accelerating}
\begin{botherref}
\oauthor{\bsnm{Gottweis}, \binits{J.}},
\oauthor{\bsnm{Weng}, \binits{W.-H.}},
\oauthor{\bsnm{Daryin}, \binits{A.}},
\oauthor{\bsnm{Tu}, \binits{T.}},
\oauthor{\bsnm{Sirkovic}, \binits{P.}},
\oauthor{\bsnm{Myaskovsky}, \binits{A.}},
\oauthor{\bsnm{Glowaty}, \binits{G.}},
\oauthor{\bsnm{Weissenberger}, \binits{F.}},
\oauthor{\bsnm{Orlandi}, \binits{A.}},
\oauthor{\bsnm{Popovici}, \binits{D.}}, et al.:
Accelerating scientific discovery with co-scientist.
Nature,
1--3
(2026)
\end{botherref}
\endbibitem

\bibitem[\protect\citeauthoryear{Ghareeb et~al.}{2026}]{ghareeb2026multi}
\begin{botherref}
\oauthor{\bsnm{Ghareeb}, \binits{A.E.}},
\oauthor{\bsnm{Chang}, \binits{B.}},
\oauthor{\bsnm{Mitchener}, \binits{L.}},
\oauthor{\bsnm{Yiu}, \binits{A.}},
\oauthor{\bsnm{Szostkiewicz}, \binits{C.J.}},
\oauthor{\bsnm{Shved}, \binits{D.}},
\oauthor{\bsnm{Gyimesi}, \binits{G.J.}},
\oauthor{\bsnm{Laurent}, \binits{J.M.}},
\oauthor{\bsnm{Wright}, \binits{S.M.}},
\oauthor{\bsnm{Razzak}, \binits{M.T.}}, et al.:
A multi-agent system for automating scientific discovery.
Nature,
1--3
(2026)
\end{botherref}
\endbibitem

\bibitem[\protect\citeauthoryear{Yang et~al.}{2026}]{endtoend2024}
\begin{botherref}
\oauthor{\bsnm{Yang}, \binits{S.}},
\oauthor{\bsnm{Chen}, \binits{F.}},
\oauthor{\bsnm{Zhao}, \binits{R.}},
\oauthor{\bsnm{Wu}, \binits{J.}},
\oauthor{\bsnm{Wang}, \binits{Y.}},
\oauthor{\bsnm{Luo}, \binits{H.}},
\oauthor{\bsnm{Han}, \binits{N.}},
\oauthor{\bsnm{Chen}, \binits{Q.}},
\oauthor{\bsnm{Hu}, \binits{Y.}},
\oauthor{\bsnm{Li}, \binits{W.}},
\oauthor{\bsnm{Li}, \binits{M.}},
\oauthor{\bsnm{Chen}, \binits{H.}},
\oauthor{\bsnm{Yang}, \binits{Y.}}:
End-to-end autonomous scientific discovery on a real optical platform.
arXiv preprint
(2026)
{\href{https://arxiv.org/abs/2604.27092}{{arXiv:2604.27092}}}
\end{botherref}
\endbibitem

\bibitem[\protect\citeauthoryear{Ghareeb et~al.}{2026}]{multiagent2024}
\begin{barticle}
\bauthor{\bsnm{Ghareeb}, \binits{A.E.}},
\bauthor{\bsnm{Chang}, \binits{B.}},
\bauthor{\bsnm{Mitchener}, \binits{L.}},
\bauthor{\bsnm{Yiu}, \binits{A.}},
\bauthor{\bsnm{Szostkiewicz}, \binits{C.J.}},
\bauthor{\bsnm{Laurent}, \binits{J.M.}},
\bauthor{\bsnm{Razzak}, \binits{M.T.}},
\bauthor{\bsnm{White}, \binits{A.D.}},
\bauthor{\bsnm{Hinks}, \binits{M.M.}},
\bauthor{\bsnm{Rodriques}, \binits{S.G.}}:
\batitle{A multi-agent system for automating scientific discovery}.
\bjtitle{Nature}
(\byear{2026})
\doiurl{10.1038/s41586-026-10652-y}
\end{barticle}
\endbibitem

\bibitem[\protect\citeauthoryear{Wang et~al.}{2024}]{wang2024agent}
\begin{barticle}
\bauthor{\bsnm{Wang}, \binits{L.}},
\bauthor{\bsnm{Ma}, \binits{C.}},
\bauthor{\bsnm{Feng}, \binits{X.}},
\bauthor{\bsnm{Zhang}, \binits{Z.}},
\bauthor{\bsnm{Yang}, \binits{H.}},
\bauthor{\bsnm{Zhang}, \binits{J.}},
\bauthor{\bsnm{Chen}, \binits{Z.}},
\bauthor{\bsnm{Tang}, \binits{J.}},
\bauthor{\bsnm{Chen}, \binits{X.}},
\bauthor{\bsnm{Lin}, \binits{Y.}}, \betal:
\batitle{A survey on large language model based autonomous agents}.
\bjtitle{Front. Comput. Sci.}
\bvolume{18}(\bissue{6}),
\bfpage{186345}
(\byear{2024})
\doiurl{10.1007/s11704-024-40231-1}
\end{barticle}
\endbibitem

\bibitem[\protect\citeauthoryear{Guo et~al.}{2024}]{li2024}
\begin{botherref}
\oauthor{\bsnm{Guo}, \binits{T.}},
\oauthor{\bsnm{Chen}, \binits{X.}},
\oauthor{\bsnm{Wang}, \binits{Y.}},
\oauthor{\bsnm{Chang}, \binits{R.}},
\oauthor{\bsnm{Pei}, \binits{S.}},
\oauthor{\bsnm{Chawla}, \binits{N.V.}},
\oauthor{\bsnm{Wiest}, \binits{O.}}:
Large language model based multi-agents: A survey of progress and challenges
(2024)
\doiurl{10.13140/RG.2.2.36311.85928}
\end{botherref}
\endbibitem

\bibitem[\protect\citeauthoryear{Gottweis et~al.}{2025}]{white2024}
\begin{botherref}
\oauthor{\bsnm{Gottweis}, \binits{J.}},
\oauthor{\bsnm{Weng}, \binits{W.-H.}},
\oauthor{\bsnm{Daryin}, \binits{A.}},
\oauthor{\bsnm{Tu}, \binits{T.}},
\oauthor{\bsnm{Palepu}, \binits{A.}},
\oauthor{\bsnm{Sirkovic}, \binits{P.}},
\oauthor{\bsnm{Myaskovsky}, \binits{A.}},
\oauthor{\bsnm{Weissenberger}, \binits{F.}},
\oauthor{\bsnm{Rong}, \binits{K.}},
\oauthor{\bsnm{Tanno}, \binits{R.}}, et al.:
Towards an {AI} co-scientist.
arXiv preprint
(2025)
{\href{https://arxiv.org/abs/2502.18864}{{arXiv:2502.18864}}}
\end{botherref}
\endbibitem

\bibitem[\protect\citeauthoryear{Boiko et~al.}{2023}]{boiko2023}
\begin{barticle}
\bauthor{\bsnm{Boiko}, \binits{D.A.}},
\bauthor{\bsnm{MacKnight}, \binits{R.}},
\bauthor{\bsnm{Kline}, \binits{B.}},
\bauthor{\bsnm{Gomes}, \binits{G.}}:
\batitle{Autonomous chemical research with large language models}.
\bjtitle{Nature}
\bvolume{624}(\bissue{7992}),
\bfpage{570}--\blpage{578}
(\byear{2023})
\doiurl{10.1038/s41586-023-06792-0}
\end{barticle}
\endbibitem

\bibitem[\protect\citeauthoryear{Szymanski et~al.}{2023}]{szymanski2023}
\begin{barticle}
\bauthor{\bsnm{Szymanski}, \binits{N.J.}},
\bauthor{\bsnm{Rendy}, \binits{B.}},
\bauthor{\bsnm{Fei}, \binits{Y.}},
\bauthor{\bsnm{Kumar}, \binits{R.E.}},
\bauthor{\bsnm{Milsted}, \binits{A.}},
\bauthor{\bsnm{McDermott}, \binits{M.}},
\bauthor{\bsnm{Gallant}, \binits{M.}},
\bauthor{\bsnm{Cubuk}, \binits{E.D.}},
\bauthor{\bsnm{Merchant}, \binits{A.}},
\bauthor{\bsnm{Kim}, \binits{H.}},
\bauthor{\bsnm{Jain}, \binits{A.}},
\bauthor{\bsnm{Bartel}, \binits{C.J.}},
\bauthor{\bsnm{Persson}, \binits{K.}},
\bauthor{\bsnm{Zeng}, \binits{Y.}},
\bauthor{\bsnm{Ceder}, \binits{G.}}:
\batitle{An autonomous laboratory for the accelerated synthesis of inorganic
  materials}.
\bjtitle{Nature}
\bvolume{624}(\bissue{7990}),
\bfpage{86}--\blpage{91}
(\byear{2023})
\doiurl{10.1038/s41586-023-06734-w}
\end{barticle}
\endbibitem

\bibitem[\protect\citeauthoryear{Lyu et~al.}{2026}]{lyu2026evoscientist}
\begin{botherref}
\oauthor{\bsnm{Lyu}, \binits{Y.}},
\oauthor{\bsnm{Zhang}, \binits{X.}},
\oauthor{\bsnm{Yi}, \binits{X.}}:
Evoscientist: towards multi-agent evolving {AI} scientists for end-to-end
  scientific discovery.
arXiv preprint
(2026)
{\href{https://arxiv.org/abs/2603.08127}{{arXiv:2603.08127}}}
\end{botherref}
\endbibitem

\bibitem[\protect\citeauthoryear{Zhang et~al.}{2026}]{zhang2026agentic}
\begin{botherref}
\oauthor{\bsnm{Zhang}, \binits{H.}},
\oauthor{\bsnm{Li}, \binits{Y.}},
\oauthor{\bsnm{Huang}, \binits{W.}},
\oauthor{\bsnm{Hou}, \binits{Z.}},
\oauthor{\bsnm{Song}, \binits{Y.}},
\oauthor{\bsnm{Liu}, \binits{X.}},
\oauthor{\bsnm{Effaty}, \binits{F.}},
\oauthor{\bsnm{Jiang}, \binits{J.}},
\oauthor{\bsnm{Wu}, \binits{S.}},
\oauthor{\bsnm{Ding}, \binits{Q.}}, et al.:
Towards agentic intelligence for materials science.
arXiv preprint
(2026)
{\href{https://arxiv.org/abs/2602.00169}{{arXiv:2602.00169}}}
\end{botherref}
\endbibitem

\bibitem[\protect\citeauthoryear{Zhu et~al.}{2026}]{zhu2026evomaster}
\begin{botherref}
\oauthor{\bsnm{Zhu}, \binits{X.}},
\oauthor{\bsnm{Cai}, \binits{Y.}},
\oauthor{\bsnm{Liu}, \binits{Z.}},
\oauthor{\bsnm{Wang}, \binits{C.}},
\oauthor{\bsnm{Li}, \binits{F.}},
\oauthor{\bsnm{Jin}, \binits{W.}},
\oauthor{\bsnm{Liu}, \binits{W.}},
\oauthor{\bsnm{Bing}, \binits{Z.}},
\oauthor{\bsnm{Zheng}, \binits{B.}},
\oauthor{\bsnm{Chai}, \binits{J.}}, et al.:
Evomaster: a foundational evolving agent framework for agentic science at
  scale.
arXiv preprint
(2026)
{\href{https://arxiv.org/abs/2604.17406}{{arXiv:2604.17406}}}
\end{botherref}
\endbibitem

\bibitem[\protect\citeauthoryear{Zhao et~al.}{2026}]{zhao2026scienceearth}
\begin{botherref}
\oauthor{\bsnm{Zhao}, \binits{Z.}},
\oauthor{\bsnm{Wen}, \binits{H.}},
\oauthor{\bsnm{Wu}, \binits{Y.}},
\oauthor{\bsnm{Ma}, \binits{J.}},
\oauthor{\bsnm{Wen}, \binits{Y.}},
\oauthor{\bsnm{Jian}, \binits{J.}},
\oauthor{\bsnm{Ge}, \binits{J.}},
\oauthor{\bsnm{Tang}, \binits{X.}},
\oauthor{\bsnm{An}, \binits{B.}},
\oauthor{\bsnm{Yin}, \binits{M.}}, et al.:
Science earth: towards a planet-scale operating system for {AI}-native
  scientific discovery.
arXiv preprint
(2026)
{\href{https://arxiv.org/abs/2606.01316}{{arXiv:2606.01316}}}
\end{botherref}
\endbibitem

\bibitem[\protect\citeauthoryear{Bianchi et~al.}{2026}]{bianchi2026collective}
\begin{botherref}
\oauthor{\bsnm{Bianchi}, \binits{F.}},
\oauthor{\bsnm{Kwon}, \binits{Y.}},
\oauthor{\bsnm{Pappu}, \binits{A.}},
\oauthor{\bsnm{Zou}, \binits{J.}}:
Harnessing the collective intelligence of {AI} agents in the wild for new
  discoveries.
arXiv preprint
(2026)
{\href{https://arxiv.org/abs/2606.10402}{{arXiv:2606.10402}}}
\end{botherref}
\endbibitem

\bibitem[\protect\citeauthoryear{Xiong et~al.}{2026}]{xiong2026evosci}
\begin{botherref}
\oauthor{\bsnm{Xiong}, \binits{X.}},
\oauthor{\bsnm{Ren}, \binits{Y.}},
\oauthor{\bsnm{Xiong}, \binits{D.}}:
Evosci: a bio-inspired multi-agent framework for the evolution of scientific
  discovery.
arXiv preprint
(2026)
{\href{https://arxiv.org/abs/2605.24018}{{arXiv:2605.24018}}}
\end{botherref}
\endbibitem

\bibitem[\protect\citeauthoryear{Bicker}{2026}]{bicker2026aster}
\begin{botherref}
\oauthor{\bsnm{Bicker}, \binits{E.}}:
Aster: autonomous scientific discovery over 20x faster than existing methods.
arXiv preprint
(2026)
{\href{https://arxiv.org/abs/2602.07040}{{arXiv:2602.07040}}}
\end{botherref}
\endbibitem

\bibitem[\protect\citeauthoryear{Liu et~al.}{2026}]{liu2026oragent}
\begin{botherref}
\oauthor{\bsnm{Liu}, \binits{Q.}},
\oauthor{\bsnm{Hao}, \binits{R.}},
\oauthor{\bsnm{Li}, \binits{C.}},
\oauthor{\bsnm{Ma}, \binits{W.}}:
Or-agent: bridging evolutionary search and structured research for automated
  algorithm discovery.
arXiv preprint
(2026)
{\href{https://arxiv.org/abs/2602.13769}{{arXiv:2602.13769}}}
\end{botherref}
\endbibitem

\bibitem[\protect\citeauthoryear{Zhang}{2026}]{zhang2026matclaw}
\begin{botherref}
\oauthor{\bsnm{Zhang}, \binits{C.}}:
Matclaw: an autonomous code-first {LLM} agent for end-to-end materials
  exploration.
arXiv preprint
(2026)
{\href{https://arxiv.org/abs/2604.02688}{{arXiv:2604.02688}}}
\end{botherref}
\endbibitem

\bibitem[\protect\citeauthoryear{Manna et~al.}{2026}]{manna2026automoose}
\begin{botherref}
\oauthor{\bsnm{Manna}, \binits{S.}},
\oauthor{\bsnm{Chan}, \binits{H.}},
\oauthor{\bsnm{Sankaranarayanan}, \binits{S.K.R.S.}}:
Automoose: an agentic {AI} for autonomous phase-field simulation.
arXiv preprint
(2026)
{\href{https://arxiv.org/abs/2603.20986}{{arXiv:2603.20986}}}
\end{botherref}
\endbibitem

\bibitem[\protect\citeauthoryear{Bisht et~al.}{2026}]{bisht2026notbuilt}
\begin{botherref}
\oauthor{\bsnm{Bisht}, \binits{H.}},
\oauthor{\bsnm{Kumar}, \binits{V.}},
\oauthor{\bsnm{Jablonka}, \binits{K.M.}},
\oauthor{\bsnm{Mausam}},
\oauthor{\bsnm{Krishnan}, \binits{N.M.A.}}:
Agentic {AI} scientists are not built for autonomous scientific discovery.
arXiv preprint
(2026)
{\href{https://arxiv.org/abs/2605.08956}{{arXiv:2605.08956}}}
\end{botherref}
\endbibitem

\bibitem[\protect\citeauthoryear{Ho et~al.}{2026}]{ho2026soundness}
\begin{botherref}
\oauthor{\bsnm{Ho}, \binits{S.T.}},
\oauthor{\bsnm{Liu}, \binits{M.}},
\oauthor{\bsnm{Nghiem}, \binits{H.}},
\oauthor{\bsnm{Huang}, \binits{F.}}:
Soundnessbench: can your {AI} scientist really tell good research ideas from
  bad ones?
arXiv preprint
(2026)
{\href{https://arxiv.org/abs/2605.30329}{{arXiv:2605.30329}}}
\end{botherref}
\endbibitem

\bibitem[\protect\citeauthoryear{Lei et~al.}{2026}]{autoresearch2026bench}
\begin{botherref}
\oauthor{\bsnm{Lei}, \binits{X.}}, et al.:
Autoresearchbench: benchmarking {AI} agents on complex scientific literature
  discovery.
arXiv preprint
(2026)
{\href{https://arxiv.org/abs/2604.25256}{{arXiv:2604.25256}}}
\end{botherref}
\endbibitem

\bibitem[\protect\citeauthoryear{Wua et~al.}{2026}]{wu2026epistemic}
\begin{botherref}
\oauthor{\bsnm{Wua}, \binits{Y.}},
\oauthor{\bsnm{Su}, \binits{T.}},
\oauthor{\bsnm{Hu}, \binits{R.}},
\oauthor{\bsnm{Zhao}, \binits{M.}},
\oauthor{\bsnm{Hu}, \binits{S.}},
\oauthor{\bsnm{Pan}, \binits{D.}},
\oauthor{\bsnm{Huang}, \binits{J.}}:
Epistemic closure: autonomous mechanism completion for physically consistent
  simulation.
arXiv preprint
(2026)
{\href{https://arxiv.org/abs/2603.09756}{{arXiv:2603.09756}}}
\end{botherref}
\endbibitem

\bibitem[\protect\citeauthoryear{Chen et~al.}{2026}]{chen2026stateful}
\begin{botherref}
\oauthor{\bsnm{Chen}, \binits{J.}},
\oauthor{\bsnm{Liu}, \binits{S.}},
\oauthor{\bsnm{Yang}, \binits{L.}}:
Statefuldiscovery: evidence-calibrated claim formation in open-ended scientific
  discovery.
arXiv preprint
(2026)
{\href{https://arxiv.org/abs/2606.11851}{{arXiv:2606.11851}}}
\end{botherref}
\endbibitem

\bibitem[\protect\citeauthoryear{Swaminathan
  et~al.}{2026}]{swaminathan2026scitrace}
\begin{botherref}
\oauthor{\bsnm{Swaminathan}, \binits{T.}},
\oauthor{\bsnm{Jiang}, \binits{R.}},
\oauthor{\bsnm{Zhang}, \binits{L.}},
\oauthor{\bsnm{Xu}, \binits{M.}}:
Scitrace: trajectory-aware safety reasoning for scientific discovery agents.
arXiv preprint
(2026)
{\href{https://arxiv.org/abs/2606.08234}{{arXiv:2606.08234}}}
\end{botherref}
\endbibitem

\bibitem[\protect\citeauthoryear{Shinn et~al.}{2023}]{shinn2023}
\begin{barticle}
\bauthor{\bsnm{Shinn}, \binits{N.}},
\bauthor{\bsnm{Cassano}, \binits{F.}},
\bauthor{\bsnm{Gopinath}, \binits{A.}},
\bauthor{\bsnm{Narasimhan}, \binits{K.}},
\bauthor{\bsnm{Yao}, \binits{S.}}:
\batitle{Reflexion: language agents with verbal reinforcement learning}.
\bjtitle{Adv. Neural Inf. Process. Syst. (NeurIPS)}
\bvolume{36},
\bfpage{8634}--\blpage{8652}
(\byear{2023})
\end{barticle}
\endbibitem

\bibitem[\protect\citeauthoryear{Yao et~al.}{2023}]{yao2023}
\begin{botherref}
\oauthor{\bsnm{Yao}, \binits{S.}},
\oauthor{\bsnm{Zhao}, \binits{J.}},
\oauthor{\bsnm{Yu}, \binits{D.}},
\oauthor{\bsnm{Du}, \binits{N.}},
\oauthor{\bsnm{Shafran}, \binits{I.}},
\oauthor{\bsnm{Narasimhan}, \binits{K.}},
\oauthor{\bsnm{Cao}, \binits{Y.}}:
React: synergizing reasoning and acting in language models.
Int. Conf. Learn. Represent. (ICLR)
(2023)
\end{botherref}
\endbibitem

\bibitem[\protect\citeauthoryear{Jia et~al.}{2026}]{jia2026scimind}
\begin{botherref}
\oauthor{\bsnm{Jia}, \binits{J.}},
\oauthor{\bsnm{Chen}, \binits{H.}},
\oauthor{\bsnm{Sun}, \binits{R.}},
\oauthor{\bsnm{Song}, \binits{Y.}},
\oauthor{\bsnm{Wang}, \binits{H.}},
\oauthor{\bsnm{Bu}, \binits{J.}},
\oauthor{\bsnm{Wu}, \binits{L.}}:
Sci-mind: cognitively-inspired adversarial debate for autonomous mathematical
  modeling.
arXiv preprint
(2026)
{\href{https://arxiv.org/abs/2603.27584}{{arXiv:2603.27584}}}
\end{botherref}
\endbibitem

\bibitem[\protect\citeauthoryear{Oh et~al.}{2026}]{oh2026multipersona}
\begin{botherref}
\oauthor{\bsnm{Oh}, \binits{J.}},
\oauthor{\bsnm{Kim}, \binits{B.}},
\oauthor{\bsnm{Li}, \binits{J.}},
\oauthor{\bsnm{Park}, \binits{Y.J.}},
\oauthor{\bsnm{Park}, \binits{J.S.}}:
Multi-persona debate system for automated scientific hypothesis generation.
arXiv preprint
(2026)
{\href{https://arxiv.org/abs/2605.23917}{{arXiv:2605.23917}}}
\end{botherref}
\endbibitem

\bibitem[\protect\citeauthoryear{Li et~al.}{2026}]{li2026causalagent}
\begin{botherref}
\oauthor{\bsnm{Li}, \binits{X.}},
\oauthor{\bsnm{Wang}, \binits{Y.}},
\oauthor{\bsnm{Li}, \binits{H.}},
\oauthor{\bsnm{Zhou}, \binits{C.}},
\oauthor{\bsnm{Gao}, \binits{E.}},
\oauthor{\bsnm{Han}, \binits{B.}},
\oauthor{\bsnm{Liu}, \binits{T.}},
\oauthor{\bsnm{Zhang}, \binits{K.}},
\oauthor{\bsnm{Bondell}, \binits{H.}},
\oauthor{\bsnm{Gong}, \binits{M.}}:
Causal ensemble agent: hierarchical causal discovery with {LLM}-guided expert
  reweighting.
arXiv preprint
(2026)
{\href{https://arxiv.org/abs/2606.10607}{{arXiv:2606.10607}}}
\end{botherref}
\endbibitem

\bibitem[\protect\citeauthoryear{Yang et~al.}{2026}]{yang2026causalab}
\begin{botherref}
\oauthor{\bsnm{Yang}, \binits{J.}},
\oauthor{\bsnm{Zhang}, \binits{D.}},
\oauthor{\bsnm{Song}, \binits{X.}},
\oauthor{\bsnm{Dai}, \binits{Q.}},
\oauthor{\bsnm{Liu}, \binits{X.}},
\oauthor{\bsnm{Chen}, \binits{Y.}},
\oauthor{\bsnm{Vashishtha}, \binits{A.}},
\oauthor{\bsnm{Shi}, \binits{J.}},
\oauthor{\bsnm{Tan}, \binits{C.}},
\oauthor{\bsnm{Peng}, \binits{H.}}:
Causalab: a scalable environment for interactive causal discovery toward {AI}
  scientists.
arXiv preprint
(2026)
{\href{https://arxiv.org/abs/2605.26029}{{arXiv:2605.26029}}}
\end{botherref}
\endbibitem

\bibitem[\protect\citeauthoryear{Vellekoop and Mosk}{2007}]{vellekoop2007}
\begin{barticle}
\bauthor{\bsnm{Vellekoop}, \binits{I.M.}},
\bauthor{\bsnm{Mosk}, \binits{A.P.}}:
\batitle{Focusing coherent light through opaque strongly scattering media}.
\bjtitle{Opt. Lett.}
\bvolume{32}(\bissue{16}),
\bfpage{2309}--\blpage{2311}
(\byear{2007})
\doiurl{10.1364/OL.32.002309}
\end{barticle}
\endbibitem

\bibitem[\protect\citeauthoryear{Popoff et~al.}{2010}]{popoff2010measuring}
\begin{barticle}
\bauthor{\bsnm{Popoff}, \binits{S.M.}},
\bauthor{\bsnm{Lerosey}, \binits{G.}},
\bauthor{\bsnm{Carminati}, \binits{R.}},
\bauthor{\bsnm{Fink}, \binits{M.}},
\bauthor{\bsnm{Boccara}, \binits{A.C.}},
\bauthor{\bsnm{Gigan}, \binits{S.}}:
\batitle{Measuring the transmission matrix in optics: An approach to the study
  and control<? format?> of light propagation in disordered media}.
\bjtitle{Physical review letters}
\bvolume{104}(\bissue{10}),
\bfpage{100601}
(\byear{2010})
\end{barticle}
\endbibitem

\bibitem[\protect\citeauthoryear{Mosk et~al.}{2012}]{mosk2012}
\begin{botherref}
\oauthor{\bsnm{Mosk}, \binits{A.P.}},
\oauthor{\bsnm{Lagendijk}, \binits{A.}},
\oauthor{\bsnm{Lerosey}, \binits{G.}},
\oauthor{\bsnm{Fink}, \binits{M.}}:
Controlling waves in space and time for imaging and focusing in complex media.
Nat. Commun.
\textbf{3}
(2012)
\doiurl{10.1038/ncomms2024}
\end{botherref}
\endbibitem

\bibitem[\protect\citeauthoryear{Rotter and Gigan}{2017}]{rotter2017light}
\begin{barticle}
\bauthor{\bsnm{Rotter}, \binits{S.}},
\bauthor{\bsnm{Gigan}, \binits{S.}}:
\batitle{Light fields in complex media: Mesoscopic scattering meets wave
  control}.
\bjtitle{Reviews of Modern Physics}
\bvolume{89}(\bissue{1}),
\bfpage{015005}
(\byear{2017})
\end{barticle}
\endbibitem

\bibitem[\protect\citeauthoryear{Bertolotti et~al.}{2012}]{bertolotti2012}
\begin{barticle}
\bauthor{\bsnm{Bertolotti}, \binits{J.}},
\bauthor{\bsnm{Putten}, \binits{E.G.}},
\bauthor{\bsnm{Blum}, \binits{C.}},
\bauthor{\bsnm{Akbulut}, \binits{D.}},
\bauthor{\bsnm{Vos}, \binits{W.L.}},
\bauthor{\bsnm{Lagendijk}, \binits{A.}},
\bauthor{\bsnm{Vos}, \binits{W.L.}}:
\batitle{Non-invasive imaging through opaque scattering layers}.
\bjtitle{Nature}
\bvolume{491}(\bissue{7423}),
\bfpage{232}--\blpage{234}
(\byear{2012})
\doiurl{10.1038/nature11578}
\end{barticle}
\endbibitem

\bibitem[\protect\citeauthoryear{Katz et~al.}{2014}]{katz2014}
\begin{barticle}
\bauthor{\bsnm{Katz}, \binits{O.}},
\bauthor{\bsnm{Heidmann}, \binits{P.}},
\bauthor{\bsnm{Fink}, \binits{M.}},
\bauthor{\bsnm{Gigan}, \binits{S.}}:
\batitle{Non-invasive single-shot imaging through scattering layers and around
  corners via speckle correlations}.
\bjtitle{Nat. Photonics}
\bvolume{8}(\bissue{10}),
\bfpage{784}--\blpage{790}
(\byear{2014})
\doiurl{10.1038/nphoton.2014.189}
\end{barticle}
\endbibitem

\bibitem[\protect\citeauthoryear{{\v{C}}i{\v{z}}m{\'a}r and
  Dholakia}{2012}]{cizmar2012}
\begin{barticle}
\bauthor{\bsnm{{\v{C}}i{\v{z}}m{\'a}r}, \binits{T.}},
\bauthor{\bsnm{Dholakia}, \binits{K.}}:
\batitle{Exploiting multimode waveguides for pure fibre-based imaging}.
\bjtitle{Nat. Commun.}
\bvolume{3},
\bfpage{1027}
(\byear{2012})
\doiurl{10.1038/ncomms2024}
\end{barticle}
\endbibitem

\bibitem[\protect\citeauthoryear{Cao et~al.}{2023}]{cao2023controlling}
\begin{barticle}
\bauthor{\bsnm{Cao}, \binits{H.}},
\bauthor{\bsnm{{\v{C}}i{\v{z}}m{\'a}r}, \binits{T.}},
\bauthor{\bsnm{Turtaev}, \binits{S.}},
\bauthor{\bsnm{Tyc}, \binits{T.}},
\bauthor{\bsnm{Rotter}, \binits{S.}}:
\batitle{Controlling light propagation in multimode fibers for imaging,
  spectroscopy, and beyond}.
\bjtitle{Advances in Optics and Photonics}
\bvolume{15}(\bissue{2}),
\bfpage{524}--\blpage{612}
(\byear{2023})
\end{barticle}
\endbibitem

\bibitem[\protect\citeauthoryear{Pl{\"o}schner
  et~al.}{2015}]{ploschner2015seeing}
\begin{barticle}
\bauthor{\bsnm{Pl{\"o}schner}, \binits{M.}},
\bauthor{\bsnm{Tyc}, \binits{T.}},
\bauthor{\bsnm{{\v{C}}i{\v{z}}m{\'a}r}, \binits{T.}}:
\batitle{Seeing through chaos in multimode fibres}.
\bjtitle{Nature photonics}
\bvolume{9}(\bissue{8}),
\bfpage{529}--\blpage{535}
(\byear{2015})
\end{barticle}
\endbibitem

\bibitem[\protect\citeauthoryear{Ouyang et~al.}{2022}]{ouyang2022}
\begin{barticle}
\bauthor{\bsnm{Ouyang}, \binits{L.}},
\bauthor{\bsnm{Wu}, \binits{J.}},
\bauthor{\bsnm{Jiang}, \binits{X.}},
\bauthor{\bsnm{Almeida}, \binits{D.}},
\bauthor{\bsnm{Wainwright}, \binits{C.L.}},
\bauthor{\bsnm{Mishkin}, \binits{P.}},
\bauthor{\bsnm{Zhang}, \binits{C.}},
\bauthor{\bsnm{Agarwal}, \binits{S.}},
\bauthor{\bsnm{Slama}, \binits{K.}},
\bauthor{\bsnm{Ray}, \binits{A.}}, \betal:
\batitle{Training language models to follow instructions with human feedback}.
\bjtitle{Adv. Neural Inf. Process. Syst. (NeurIPS)}
\bvolume{35},
\bfpage{27730}--\blpage{27744}
(\byear{2022})
\end{barticle}
\endbibitem

\bibitem[\protect\citeauthoryear{Wei et~al.}{2022}]{wei2022}
\begin{barticle}
\bauthor{\bsnm{Wei}, \binits{J.}},
\bauthor{\bsnm{Wang}, \binits{X.}},
\bauthor{\bsnm{Schuurmans}, \binits{D.}},
\bauthor{\bsnm{Bosma}, \binits{M.}},
\bauthor{\bsnm{Ichter}, \binits{B.}},
\bauthor{\bsnm{Xia}, \binits{F.}},
\bauthor{\bsnm{Chi}, \binits{E.H.}},
\bauthor{\bsnm{Le}, \binits{Q.V.}},
\bauthor{\bsnm{Zhou}, \binits{D.}}:
\batitle{Chain-of-thought prompting elicits reasoning in large language
  models}.
\bjtitle{Adv. Neural Inf. Process. Syst. (NeurIPS)}
\bvolume{35},
\bfpage{24824}--\blpage{24837}
(\byear{2022})
\end{barticle}
\endbibitem

\bibitem[\protect\citeauthoryear{Schick et~al.}{2023}]{schick2023}
\begin{barticle}
\bauthor{\bsnm{Schick}, \binits{T.}},
\bauthor{\bsnm{Dwivedi-Yu}, \binits{J.}},
\bauthor{\bsnm{Dess{\`i}}, \binits{R.}},
\bauthor{\bsnm{Raileanu}, \binits{R.}},
\bauthor{\bsnm{Lomeli}, \binits{M.}},
\bauthor{\bsnm{Zettlemoyer}, \binits{L.}},
\bauthor{\bsnm{Cancedda}, \binits{N.}},
\bauthor{\bsnm{Scialom}, \binits{T.}}:
\batitle{Toolformer: language models can teach themselves to use tools}.
\bjtitle{Adv. Neural Inf. Process. Syst. (NeurIPS)}
\bvolume{36},
\bfpage{68539}--\blpage{68551}
(\byear{2023})
\end{barticle}
\endbibitem

\bibitem[\protect\citeauthoryear{Sumers et~al.}{2023}]{sumers2024}
\begin{botherref}
\oauthor{\bsnm{Sumers}, \binits{T.R.}},
\oauthor{\bsnm{Yao}, \binits{S.}},
\oauthor{\bsnm{Narasimhan}, \binits{K.}},
\oauthor{\bsnm{Griffiths}, \binits{T.L.}}:
Cognitive architectures for language agents.
arXiv preprint
(2023)
{\href{https://arxiv.org/abs/2309.02427}{{arXiv:2309.02427}}}
\end{botherref}
\endbibitem

\bibitem[\protect\citeauthoryear{Zhu et~al.}{2026}]{zhu2026lap}
\begin{botherref}
\oauthor{\bsnm{Zhu}, \binits{L.}},
\oauthor{\bsnm{Gao}, \binits{L.}},
\oauthor{\bsnm{Chen}, \binits{Y.}},
\oauthor{\bsnm{Zhu}, \binits{D.}},
\oauthor{\bsnm{Huang}, \binits{J.}}:
{LAP}: an agent-to-instrument protocol for autonomous science.
arXiv preprint
(2026)
{\href{https://arxiv.org/abs/2606.03755}{{arXiv:2606.03755}}}
\end{botherref}
\endbibitem

\bibitem[\protect\citeauthoryear{Yang et~al.}{2024}]{chen2022}
\begin{barticle}
\bauthor{\bsnm{Yang}, \binits{X.}}, \betal:
\batitle{Curriculum learning for ab initio deep learned refractive optics}.
\bjtitle{Nature Communications}
\bvolume{15}(\bissue{1}),
\bfpage{6572}
(\byear{2024})
\doiurl{10.1038/s41467-024-50835-7}
\end{barticle}
\endbibitem

\bibitem[\protect\citeauthoryear{Zeng et~al.}{2026}]{zeng2026physics}
\begin{botherref}
\oauthor{\bsnm{Zeng}, \binits{X.}},
\oauthor{\bsnm{Zang}, \binits{Y.}},
\oauthor{\bsnm{Liu}, \binits{P.}},
\oauthor{\bsnm{Yu}, \binits{F.}},
\oauthor{\bsnm{Yang}, \binits{Y.}},
\oauthor{\bsnm{{\v{C}}i{\v{z}}m{\'a}r}, \binits{T.}},
\oauthor{\bsnm{Du}, \binits{Y.}}:
Physics-guided foundation model for universal speckle removal in ultrathin
  multimode fiber imaging.
arXiv preprint arXiv:2601.06448
(2026)
\end{botherref}
\endbibitem

\bibitem[\protect\citeauthoryear{Xu et~al.}{2025}]{Xu2025Dual}
\begin{barticle}
\bauthor{\bsnm{Xu}, \binits{J.}}, \betal:
\batitle{Dual holographic and polarization encoding for high fidelity image
  transmission through multimode fibers}.
\bjtitle{Optics \& Laser Technology}
\bvolume{191},
\bfpage{113301}
(\byear{2025})
\doiurl{https://doi.org}
\end{barticle}
\endbibitem

\bibitem[\protect\citeauthoryear{Ronneberger et~al.}{2015}]{ronneberger2015}
\begin{botherref}
\oauthor{\bsnm{Ronneberger}, \binits{O.}},
\oauthor{\bsnm{Fischer}, \binits{P.}},
\oauthor{\bsnm{Brox}, \binits{T.}}:
U-net: convolutional networks for biomedical image segmentation.
Med. Image Comput. Comput.-Assist. Interv. (MICCAI),
234--241
(2015)
\doiurl{10.1007/978-3-319-24574-4_28}
\end{botherref}
\endbibitem

\bibitem[\protect\citeauthoryear{Cui}{2026}]{cui2026inquitree}
\begin{botherref}
\oauthor{\bsnm{Cui}, \binits{S.}}:
Inquitree: evaluating {AI} agents in the scientific inquiry loop with
  paper-derived research trees.
arXiv preprint
(2026)
{\href{https://arxiv.org/abs/2606.09550}{{arXiv:2606.09550}}}
\end{botherref}
\endbibitem

\bibitem[\protect\citeauthoryear{Stach et~al.}{2021}]{raj2024}
\begin{barticle}
\bauthor{\bsnm{Stach}, \binits{E.}}, \betal:
\batitle{Autonomous experimentation systems for materials development: A
  community perspective}.
\bjtitle{Matter}
\bvolume{4}(\bissue{9}),
\bfpage{2702}--\blpage{2726}
(\byear{2021})
\doiurl{10.1016/j.matt.2021.06.036}
\end{barticle}
\endbibitem

\bibitem[\protect\citeauthoryear{Song et~al.}{2026}]{song2026catalyst}
\begin{botherref}
\oauthor{\bsnm{Song}, \binits{Z.}},
\oauthor{\bsnm{Zhang}, \binits{Z.}},
\oauthor{\bsnm{Cheng}, \binits{L.}}:
Autonomous heterogeneous catalyst discovery with a self-evolving multi-agent
  digital twin.
arXiv preprint
(2026)
{\href{https://arxiv.org/abs/2606.05050}{{arXiv:2606.05050}}}
\end{botherref}
\endbibitem

\bibitem[\protect\citeauthoryear{Jiang et~al.}{2026}]{jiang2026deltaevolve}
\begin{botherref}
\oauthor{\bsnm{Jiang}, \binits{J.}},
\oauthor{\bsnm{Ding}, \binits{T.}},
\oauthor{\bsnm{Zhu}, \binits{Z.}}:
Deltaevolve: accelerating scientific discovery through momentum-driven
  evolution.
arXiv preprint
(2026)
{\href{https://arxiv.org/abs/2602.02919}{{arXiv:2602.02919}}}
\end{botherref}
\endbibitem

\bibitem[\protect\citeauthoryear{Krenn et~al.}{2022}]{krenn2022}
\begin{barticle}
\bauthor{\bsnm{Krenn}, \binits{M.}},
\bauthor{\bsnm{Pollice}, \binits{R.}},
\bauthor{\bsnm{Guo}, \binits{S.Y.}},
\bauthor{\bsnm{Aldeghi}, \binits{M.}},
\bauthor{\bsnm{Cervera-Lierta}, \binits{A.}},
\bauthor{\bsnm{Friederich}, \binits{P.}},
\bauthor{\bsnm{Passos~Gomes}, \binits{G.}},
\bauthor{\bsnm{H{\"a}se}, \binits{F.}},
\bauthor{\bsnm{Jinich}, \binits{A.}},
\bauthor{\bsnm{Nigam}, \binits{A.}}, \betal:
\batitle{On scientific understanding with artificial intelligence}.
\bjtitle{Nat. Rev. Phys.}
\bvolume{4}(\bissue{12}),
\bfpage{761}--\blpage{769}
(\byear{2022})
\doiurl{10.1038/s42254-022-00518-3}
\end{barticle}
\endbibitem

\bibitem[\protect\citeauthoryear{Lin et~al.}{2026}]{lin2026compphysics}
\begin{botherref}
\oauthor{\bsnm{Lin}, \binits{H.}},
\oauthor{\bsnm{Liu}, \binits{C.}},
\oauthor{\bsnm{Yan}, \binits{G.}}:
Large language model based agent for automated discovery in computational
  physics.
arXiv preprint
(2026)
{\href{https://arxiv.org/abs/2606.14266}{{arXiv:2606.14266}}}
\end{botherref}
\endbibitem

\bibitem[\protect\citeauthoryear{Somasekharan
  et~al.}{2026}]{somasekharan2026aicfd}
\begin{botherref}
\oauthor{\bsnm{Somasekharan}, \binits{N.}},
\oauthor{\bsnm{Pathak}, \binits{R.}},
\oauthor{\bsnm{Dhanakoti}, \binits{M.}},
\oauthor{\bsnm{Zhang}, \binits{T.}},
\oauthor{\bsnm{Yue}, \binits{L.}},
\oauthor{\bsnm{Zhu}, \binits{A.}},
\oauthor{\bsnm{Pan}, \binits{S.}}:
{AI} {CFD} scientist: toward open-ended computational fluid dynamics discovery
  with physics-aware {AI} agents.
arXiv preprint
(2026)
{\href{https://arxiv.org/abs/2605.06607}{{arXiv:2605.06607}}}
\end{botherref}
\endbibitem

\bibitem[\protect\citeauthoryear{He et~al.}{2016}]{he2016}
\begin{botherref}
\oauthor{\bsnm{He}, \binits{K.}},
\oauthor{\bsnm{Zhang}, \binits{X.}},
\oauthor{\bsnm{Ren}, \binits{S.}},
\oauthor{\bsnm{Sun}, \binits{J.}}:
Deep residual learning for image recognition.
Proc. IEEE Conf. Comput. Vis. Pattern Recognit. (CVPR),
770--778
(2016)
\doiurl{10.1109/CVPR.2016.90}
\end{botherref}
\endbibitem

\bibitem[\protect\citeauthoryear{Vaswani et~al.}{2017}]{vaswani2017}
\begin{barticle}
\bauthor{\bsnm{Vaswani}, \binits{A.}},
\bauthor{\bsnm{Shazeer}, \binits{N.}},
\bauthor{\bsnm{Parmar}, \binits{N.}},
\bauthor{\bsnm{Uszkoreit}, \binits{J.}},
\bauthor{\bsnm{Jones}, \binits{L.}},
\bauthor{\bsnm{Gomez}, \binits{A.N.}},
\bauthor{\bsnm{Kaiser}, \binits{{\L}.}},
\bauthor{\bsnm{Polosukhin}, \binits{I.}}:
\batitle{Attention is all you need}.
\bjtitle{Adv. Neural Inf. Process. Syst. (NeurIPS)}
\bvolume{30},
\bfpage{5998}--\blpage{6008}
(\byear{2017})
\end{barticle}
\endbibitem

\end{thebibliography}

\end{document}